\documentclass{ieeetj}

\usepackage{cite}
\usepackage{amsmath,amssymb,amsfonts}
\usepackage{graphicx,color}
\usepackage{textcomp}
\usepackage{xcolor}
\usepackage{xspace}
\usepackage{hyperref}
\hypersetup{hidelinks}
\usepackage{booktabs}
\usepackage{tikz}
\usepackage{pgfplots}
\usetikzlibrary{positioning, shapes, arrows, automata}
\usepackage[normalem]{ulem}

\usepackage{multirow}

\usepackage[ruled,lined]{algorithm2e}

\usepackage{algpseudocode}
\usepackage{glossaries}
\let\labelindent\relax
\usepackage[inline]{enumitem}
\pgfplotsset{compat=1.14}
\usepackage[T1]{fontenc}
\usepackage{lmodern}
\def\BibTeX{{\rm B\kern-.05em{\sc i\kern-.025em b}\kern-.08em
    T\kern-.1667em\lower.7ex\hbox{E}\kern-.125emX}}
\AtBeginDocument{\definecolor{tmlcncolor}{cmyk}{0.93,0.59,0.15,0.02}\definecolor{NavyBlue}{RGB}{0,86,125}}

\newcommand{\eg}{\textit{e.g., }}

\newcommand{\etal}{et al. }
\newacronym{cots}{COTS}{commercial-off-the-shelf}
\newacronym{uav}{UAV}{unmanned aerial vehicle}
\newacronym{ugv}{UGV}{unmanned ground vehicle}
\newacronym{grd}{GRD}{ground resolution distance}
\newacronym{sfm}{SFM}{structure from motion}
\newacronym{hdr}{HDR}{high dynamic range}
\newacronym{fov}{FoV}{field of view}
\newacronym{agl}{AGL}{above ground level}
\newacronym{psnr}{PSNR}{peak signal-to-noise ratio}
\newacronym{ssim}{SSIM}{structural similarity index}
\newacronym{imu}{IMU}{inertial measurement unit}
\newacronym{odm}{ODM}{OpenDroneMap}
\newacronym{utm}{UTM}{Universal Transverse Mercator}
\newacronym{los}{LOS}{line-of-sight}
\newacronym{rssi}{RSSI}{received signal strength indicator}
\newacronym{swap}{SWaP}{size, weight, and power}
\newacronym{gnss}{GNSS}{Global Navigation Satellite System}

\newif\ifdraftcolor

\ifdraftcolor
\newcommand{\zac}[1]{{\color{orange}(Zac) #1}}
\newcommand{\vm}[1]{{\color{teal}(Varun) #1}}
\newcommand{\fclad}[1]{{\color{olive}(Fclad) #1}}
\newcommand{\jh}[1]{{\color{violet}(Jason) #1}}

\newcommand{\deletetext}[1]{{\color{red}\sout{#1}}}

\newcommand{\edcomment}[1]{{\color{cyan}#1}}

\else
\newcommand{\zac}[1]{}
\newcommand{\vm}[1]{}
\newcommand{\fclad}[1]{}
\newcommand{\jh}[1]{}

\newcommand{\deletetext}[1]{}
\newcommand{\edcomment}[1]{}
\fi 

\def\OJlogo{}
\def\seclogo{}

\def\authorrefmark#1{\ensuremath{^{\textbf{#1}}}}

\begin{document}

\newcommand{\papertitle}[0]{Air-Ground Collaboration for Language-Specified Missions in Unknown Environments}
\markboth{\papertitle}{Cladera*, Ravichandran*, Hughes* {et al.}}

\title{\papertitle}

\author{Fernando Cladera\authorrefmark{*1}, Zachary Ravichandran\authorrefmark{*1}, Jason Hughes\authorrefmark{*1}, Varun Murali\authorrefmark{1}, Carlos Nieto-Granda\authorrefmark{2}, M. Ani Hsieh\authorrefmark{1}, George J. Pappas\authorrefmark{1}, Camillo J. Taylor\authorrefmark{1}, and Vijay Kumar\authorrefmark{1}
\affil{GRASP Laboratory, University of Pennsylvania, Philadelphia, Pennsylvania}
\affil{U.S. DEVCOM Army Research Laboratory (ARL), Adelphi, Maryland\\ $^*$Equal Contribution}
\corresp{Corresponding authors: Fernando Cladera (fclad@seas.upenn.edu), Zachary Ravichandran (zacravi@seas.upenn.edu), \\and Jason Hughes (jasonah@seas.upenn.edu).}
\authornote{We gratefully acknowledge the support of
ARL DCIST CRA W911NF-17-2-0181,
NIFA grant 2022-67021-36856,
the IoT4Ag Engineering Research Center funded by the National Science Foundation (NSF) under NSF Cooperative Agreement Number EEC-1941529, 
NVIDIA, %
and the NSF Graduation Research Fellowship Program.}
}

\begin{abstract}
As autonomous robotic systems become increasingly mature, users will want to specify missions at the level of intent rather than in low-level detail.
Language is an expressive and intuitive medium for such mission specification.
However, realizing language-guided robotic teams requires overcoming significant technical hurdles.
Interpreting and realizing language-specified missions requires advanced semantic reasoning.
Successful heterogeneous robots must effectively coordinate actions and share information across varying viewpoints.
Additionally, communication between robots is typically intermittent, necessitating robust strategies that leverage communication opportunities to maintain coordination and achieve mission objectives.
In this work, we present a first-of-its-kind system where an \gls{uav} and an \gls{ugv} are able to collaboratively accomplish missions specified in natural language while reacting to changes in specification on the fly.
We leverage a Large Language Model (LLM)-enabled planner to reason over semantic-metric maps that are built online and opportunistically shared between an aerial and a ground robot.
We consider task-driven navigation in urban and rural areas. Our system must infer mission-relevant semantics and actively acquire information via semantic mapping.
In both ground and air-ground teaming experiments, we demonstrate our system on seven different natural-language specifications at up to kilometer-scale navigation.

\end{abstract}



\maketitle
\begin{figure*}[t!]
    \centering
    \includegraphics[width=0.9\textwidth]{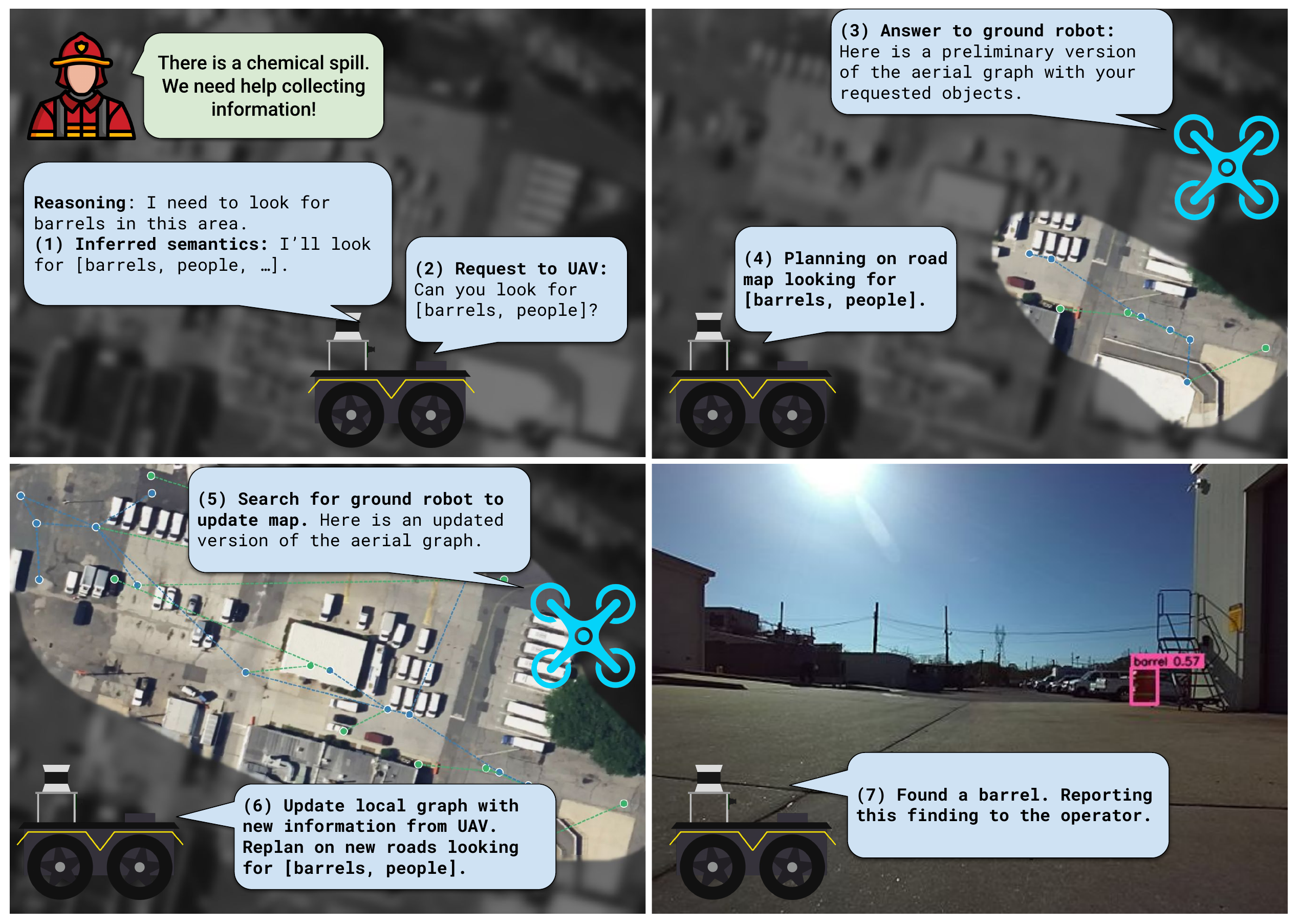}
    \caption{Example mission specification. The \gls{uav} and \glspl{ugv} start their mission in the same location. The \glspl{ugv} can task the \gls{uav} with a list of labels to look for. Once it receives a graph, the \glspl{ugv} can plan on the \gls{uav} map, looking for objects of interest. Satellite image in the background is only provided for representation purposes.}
    \label{fig:mission_concept}
    \vspace{-.3cm} 
\end{figure*}

\section{INTRODUCTION}
Exploration of unknown unstructured environments is one of the quintessential field robotics applications, with applications to infrastructure inspection \cite{lattanzi2017review}, search and rescue \cite{chung2023into}, disaster response\cite{kruijff2012rescue}, law enforcement \cite{nguyen2001robotics}, and crop inspection\cite{fountas2020agricultural}, among others.
Heterogeneous teams of aerial and ground robots have a distinct advantage for fast exploration missions, by providing complementary capabilities compared to a team of identical robots.

For example, \glspl{uav} flying at high altitudes move in an obstacle-free space; thus, they can achieve higher speeds and
provide an elevated vantage point.
\glspl{uav} can trade off resolution for altitude and fly higher when a fast scene overview is required.
Moreover, high-altitude \glspl{uav} can act as obstacle-free communication nodes, achieving line-of-sight with other nodes on the ground.
Unfortunately, the low \gls{swap}  constraints on \glspl{uav} restrict the payload, compute, and operation time of \glspl{uav}.
Contrarily,
\glspl{ugv} are not affected by these power limitations: they can carry heavier payloads and operate for extended periods.
However, \glspl{ugv} speed is limited as they have to plan trajectories to avoid static and dynamic obstacles and consider the traversability of the terrain where they operate.

Liu et al.~\cite{liu2022review} identified different roles \glspl{uav} and \glspl{ugv} perform in a team, such as sensor, actuator or auxiliary. \textbf{We focus on applications where \glspl{uav} and \glspl{ugv} act cooperatively to perform a mission}.
However, in most examples of the literature \cite{liu2022review}, the tasks are pre-defined for the robots. For instance, \glspl{uav} may be used for to localize  ground robots~\cite{chaimowicz2004experiments} or generate maps and traversability information~\cite{miller2022stronger}. Miller et al.~\cite{spomp_journal} were one of the first to use \glspl{uav} in a multi-role setting, where the \gls{uav} generated a map and acted as an active communication relay~\cite{cladera2024enabling}.

Still, the state of the art is far from a seamless collaboration of aerial and ground robots, where tasks are not pre-encoded.
For example, leading systems in the DARPA SubT challenge, including Nebula~\cite{agha2021nebula} and Cerberus~\cite{tranzatto2022cerberus}, were designed for object search over a specific semantic set.
Recent literature proposes planners for language-specified tasks in single-robot, small-scale, structured environments (\eg indoors)~\cite{saycan2022arxiv,werby23hovsg,liu23lang2ltl}.
However, such work assumes favorable conditions, including near-perfect environment knowledge and perfect communications~\cite{rana2023sayplan}.

As autonomous robotic systems become increasingly mature, we envision a heterogeneous robotic team that users can task at the level of intent rather than detailed and step-by-step instructions.
Given a user's mission specification, the robot team would infer the required subtasks, reason about each platform's constraints and abilities, and execute a plan suited to the environment of operation.
Realizing this vision on robotic systems, and particularly air-ground robotics, remains a significant challenge.
Existing literature generally assumes mission specifications are fixed.
Creating planners that reason over dynamic mission specifications, where no task is pre-specified on either the aerial or ground vehicle, remains an open research question.
Correspondingly, existing literature fix semantics upfront or only share metric information among robots.
However, dynamically configuring semantics is a desirable property.
Finally, robot mobility should not be limited by the communication network infrastructure, and robots should not need be in continuous communication range.
Creating robot teams that leverage fully opportunistic communications is an ongoing challenge.

This work builds upon our air-ground collaboration work~\cite{miller2022stronger, spomp_journal}, including components of opportunistic communication~\cite{cladera2024enabling} and language-driven planning~\cite{ravichandran_jackal}.
We aim to address the challenges posed in~\cite{cladera2024challenges} by proposing a language-driven air-ground teaming system with the following \textbf{novel contributions}:

\begin{itemize}
    \item The \emph{first} air-ground teaming system for language-specified missions in  unknown environments.
    \item A \emph{decentralized and compact} semantic mapping approach that enables the \gls{uav} and \glspl{ugv} to share observations  and \emph{dynamically} set mission-relevant semantics of interest to map.

\end{itemize}
We performed experimental validation and system deployment in kilometer-scale experiments in semi-urban and rural environments.
An example mission scenario is shown in Fig.~\ref{fig:mission_concept}.
The supplementary material for this work is available online\footnote{\url{http://tfr-air-ground.fcladera.com}}.

\begin{figure*}[ht!]
    \centering
    \includegraphics[width=\textwidth]{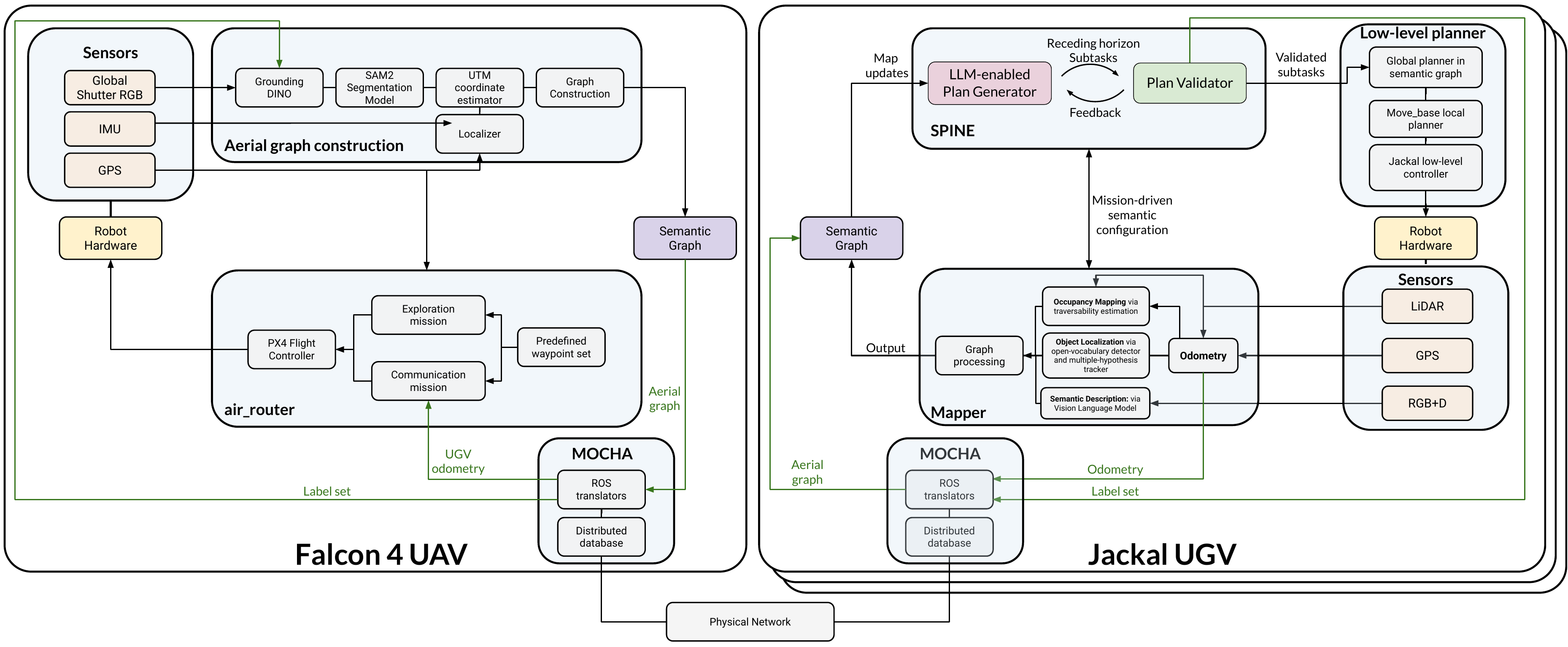}
    \caption{System overview. Green lines represent information that is transmitted to and from MOCHA and thus are communicated to other robots in the system.}
    \label{fig:sys-overview}
    \vspace{-.2cm} 
\end{figure*}

\section{RELATED WORK}
\label{sec:related_work}
\subsection{Air-ground Robot Teaming}

Cooperative air-ground collaboration in robotics has more than twenty years of development~\cite{liu2022review}.
The initial works of Elfes et al. used robotics blimps to transmit aerial images to ground robots, which used them for visual navigation~\cite{elfes1999air}.
Other early works also use aerial vehicles for localization, tracking, and obstacle avoidance~\cite{chaimowicz2004experiments, stentz2003real, hsieh2007adaptive}.
In these works, the \gls{uav} acts as an \emph{eye in the sky}, providing information that compliments ground robot knowledge.
One of the first examples of air-ground mapping is described in~\cite{kruijff2012rescue} and~\cite{michael2014collaborative}, where a team of robots was deployed to assess the damages of historical buildings after earthquakes in Italy and Japan.
These works proved the potential of air-ground teams in disaster scenarios by reducing the risk of rescue personnel.

Most existing air-ground robot teaming literature focuses on static task assignments, where the robots' tasks are pre-agreed.
For example, the \gls{uav} provides aerial information for mapping or obstacle avoidance.
In some works~\cite{cladera2024enabling, spomp_journal}, the UAV has multiple roles as a map provider and communication relay.
Still, few works show an explicit tasking of the \gls{uav}-\gls{ugv} team.
\textbf{In this work, we aim to address this issue by allowing the \gls{ugv} to task the \gls{uav} towards an integrated semantic exploration task}.

\subsection{Semantic Representations for Navigation}
Semantic planning requires representations that contain the contextual information for high-level planning and traversability information for lower-level control.
The perception community has advanced online semantic planning for tasks such as active exploration and object search~\cite{asgharivaskasi2023semantic, active3dsemslam, spomp_journal, kurtz2024real}.
Most of the semantic mapping literature focuses on single-robot applications.
In these works, scene graphs have emerged as a popular representation for semantic planning, as they capture objects, topology, and traversable regions.
Hydra provides a real-time scene graph engine~\cite{hughes2024foundations} designed for indoor environments.
Strader \etal~\cite{outdoor_dsg} relax this assumption.
Topological maps are similar but do not include a hierarchy~\cite{RoboHop, chiang2024mobilityvlamultimodalinstruction}.
Recent work incorporates foundation models into mapping pipelines to create open-vocabulary representations.
Mappers, including ConceptGraphs~\cite{conceptgraphs}, HOVSG~\cite{werby23hovsg}, and Clio~\cite{Maggio2024Clio}, assign semantic feature vectors to entities in the map. Semantic labels are then produced at runtime, depending on the task.

Effective representations for multi-robot teaming share many of the aforementioned properties.
One of the major additional challenges is that they require a consistent frame between robots.
Consistent multi-robot state estimation is addressed by works such as Kimera-multi~\cite{tian22tro_kimeramulti}.
Similarly, Hydra-Multi builds scene graphs across a robot team via loop-closure detection and relative state estimation \cite{chang2023hydra}.
Alternatively, the SPOMP system addresses this problem with a semantics-based relative localization module~\cite{spomp_journal}.
However, this approach requires an orthomap to be constructed in real time and transmitted to the ground robots.
Sparsity is another desirable property, as communication and compute may be limited, which this work addresses.

\textbf{This work introduces a sparse map representation shared by the \gls{uav} and \gls{ugv}}. We demonstrate how the \gls{uav} can build the map in real-time and how the \gls{ugv} can use such representation to execute its mission while augmenting it with newly observed information.

\subsection{Language-specified mission planning}
Motivated by the increasing maturity and accessibility of LLMs, the robotics community has studied the use of language for specifying tasks and missions.
LLM-enabled planners have been developed for mobile manipulation~\cite{rana2023sayplan, saycan2022arxiv, momallm24}, service robotics~\cite{llm_service_robot}, autonomous driving~\cite{sharan2023llm}, navigation~\cite{xie2023reasoning, pmlr-v229-shah23c, pmlr-v205-shah23b, vlmaps}, and fault detection~\cite{SinhaElhafsiEtAl2024Aesop, tagliabue2023real}
These methods typically configure a pre-trained LLM, such as GPT-4~\cite{openai2024gpt4technicalreport}, via in-context prompts.
This approach avoids fine-tuning or retraining an LLM, which is computationally expensive, but it still channels the LLM's common sense into a specific problem~\cite{huang2022inner}.
The LLM-enabled planner is then provided a set of behaviors such as graph navigation goals~\cite{sharan2023llm}, predicates in a formal planning language~\cite{liu23lang2ltl}, lower-level APIs for code generation~\cite{codeaspolicies2022, ma2023eureka, ma2024dreureka}, or learned behaviors~\cite{saycan2022arxiv}.
At runtime, the LLM is given an environment map, such as a graph~\cite{rana2023sayplan} or semantic regions~\cite{huang2022inner}.
A line of research plans over formal languages such as Linear Temporal Logic (LTL) or Planning Domain Definition Language (PDDL)
~\cite{dai2024optimalscenegraphplanning, liu23lang2ltl, pmlr-v229-liu23d, chen2023nl2tl, chen2023autotamp, garg2024largelanguagemodelsrescue, liu2023llmp}.
Recent literature also considers language-specified missions for multi-robot systems~\cite{chen2024scalable, kannan2023smart,roco_multi_llm, wang2024safe, liu2024coherent}.
However, these works consider controlled simulation or closed-world manipulation environments with perfect or near-perfect environment knowledge.

In the above approaches, instructions typically state the required subtasks and required semantics~\cite{verify_llm_ltl}.
While works such as SayPlan consider less well-specified tasks (``find me something to drink''), they still assume pre-mapped or highly-structured indoor scenes~\cite{rana2023sayplan, momallm24}.
Other research relaxes the requirement of a pre-built semantic map by incorporating feedback from perception systems~\cite{huang2022inner, momallm24, SinhaElhafsiEtAl2024Aesop} or specifying semantics at runtime~\cite{verify_llm_ltl, chen2022nlmapsaycan}.
However, perception is limited to object detection or designed for small room-centric environments where the planner can leverage clear hierarchy and natural bounds on the environment.

We recently introduced the SPINE planning architecture, which serves as the base of our UGV planner~\cite{ravichandran_jackal}.
In contrast to the above works, SPINE reasons over under-specified missions where the user provides high-level intent.
Furthermore, the environment is only partially-known or unknown at runtime, so the system must actively acquire task-relevant information.
Compared to our previous work, we \textbf{extend SPINE to enable tasking the UAV}, allowing it to generate a better prior for the \gls{ugv} mission.

\subsection{Multi-robot communications}
Communication in air-ground teams can be broadly grouped into three groups: a) ensured network connectivity, b) rendezvous approaches, and c) opportunistic communication. Ensured network communication approaches focus on deploying a group of robots such that there is guaranteed communication between all network nodes~\cite{hsieh2008maintaining, stump2011visibility}.
Continuous communication approaches are key when a human operator is required in the loop or when robots need to
report to a centralized base station continuously.
The main disadvantage of this approach is that the experiment area is limited by the communication channel quality, and it can be severely limited in environments with non-\gls{los} between robots.

Rendezvous approaches rely on established guarantees for communication as robots physically meet to exchange information~\cite{roy2001collaborative, xi2020syntesis}.
Alternatively, robots can form continuous communication networks and venture out to areas without communications~\cite{saboia2022achord, Ginting2021CHORDDD}, with successful application in underground mapping at the SubT challenge.
Still, these techniques limit
mobility when the number of robots is reduced.

In~\cite{cladera2024enabling, spomp_journal}, we introduced MOCHA, a fully opportunistic communication approach for air-ground teams.
MOCHA is based on a gossip communication approach~\cite{birman2007promise}, that enables large-scale exploration, as communication constraints do not limit robot operations.
To avoid robots in isolated situations, the aerial robot acts as a \emph{data mule}, finding ground robots and physically carrying data from them.
We showed that MOCHA performs well in large-scale environments, in simulation and
real-world experiments.
\textbf{This work leverages MOCHA for communication between the robots, and for the communication-aware planner of the \gls{uav}}.

\subsection{Robot teaming with human in the loop}
Human-in-the-loop frameworks have become pivotal in improving robot teaming by integrating human oversight and domain-specific tasking into multi-robot systems.
Measuring the efficacy of human-robot teaming has been studied extensively in \cite{novitzky2021toward, kaupp2008measuring, ma2022metrics}.
Specifically, Novitzky et al. uses the game ``capture the flag'' in \cite{novitzky2021toward} and \cite{novitzky2019aquaticus} to assess effective ways to task robot teammates.
Robot teams with humans in-the-loop have been developed in \cite{fink2016airground, miller2020subt, hughes2023collaborative} for various
inspection tasks.
All of these rely on predefining what is to be inspected before the system is running by hard-coding in \cite{hughes2023collaborative} or predefining semantic labels as in \cite{miller2020subt}.
All of these systems lack the ability for the human to re-task the robot team on the fly like our system can.
Prior work has used different interfaces to interact with the robot team, including map interfaces \cite{larkin2021atak} that allow for \texttt{goto} commands or \texttt{search} commands,  and virtual reality \cite{kim2022aerial} and gesture control \cite{lavanya2017gesture}, which allow for advanced teleoperation of the robot.
These interfaces require more input from the human-in-the-loop, which means the robot system is making fewer decisions.
All of these interfaces fail to be as expressive as natural language for the human as our implementation of large language models in a human-robot team.
\textbf{In comparison, we present the \emph{first} heterogeneous teaming system with a human in the loop capable of interacting through natural language.}

\section{SYSTEM OVERVIEW}
\label{sec:sys_overview}
\begin{table}[t]
    \centering
    \begin{tabular}{c c c c}
        \toprule
        Producer & Name & Message type & Size \\
        \toprule
          \gls{ugv} & \texttt{UGV\_odometry} & Pose stamped & 74 B\\
          \gls{ugv} & \texttt{label\_set} & String (JSON) & 2B\\
          \gls{uav} & \texttt{aerial\_graph} & String (JSON) & 40.2 KB \\
          \toprule
    \end{tabular}
    \caption{Messages transmitted between robots using MOCHA, our opportunistic communications framework, after 10 minutes of an experiment.}
    \vspace{-.6cm} 
    \label{tab:mochamessages}
\end{table}

\subsection{Mission specification}
This section describes the overall system architecture and mission specification using language as depicted in Fig. \ref{fig:mission_concept}.
A human operator requests assistance with a particular task from one or more \glspl{ugv}, which infers the semantics in the scene that need to be found.
The \glspl{ugv} also infers a list of labels that it will request the \gls{uav} to search for, in addition to traversable classes, like paved roads.

The \gls{uav} then produces an initial graph with traversable regions and objects of interest, which is incrementally transmitted through MOCHA to the \glspl{ugv}.
New labels can be added on-demand by the \gls{ugv} and sent to the \gls{uav}.
The \gls{uav} also acts as a communication relay, carrying messages between the different robots and the operators if needed.

Once a graph has been received, the \glspl{ugv} plan a trajectory using the graph to
objects of interest.
If objects are found, they are reported to the operator.
Our current system can run as many ground robots as required, but no coordination occurs between them.
Additionally, we can pre-generate a partial environment graph using other data sources, such as satellite images.

\subsection{System architecture}

A system overview is presented in Fig.~\ref{fig:sys-overview}.
The different sub-components are described in Sec.~\ref{sec:methodology}.
We use a \gls{gnss} to establish the positions of the different robots, as well as to annotate the coordinates of the nodes in the graph.

\subsection{Communications}
\label{subsec:communications}
The list of messages transmitted by each robot over MOCHA is described in Tab.~\ref{tab:mochamessages}, with a representative size for each compressed message 10 minutes after the start of the mission.

We used Rajant~\cite{rajantBreadcrumbWireless} breadcrumb radios for our physical communication layer which MOCHA runs on top of.
Robots use the Rajant DX-2 and Cardinal radios, whereas the base station runs a FE1 radio.
The Rajant Breadcrumb API is used to obtain information regarding the link quality, such as the \gls{rssi}.
This information is used to trigger a communication exchange between two robots.

We also deployed \emph{dummy} ME4 nodes to act as communication relays for LLM API calls for SPINE.
These nodes do not participate actively in the opportunistic communication process.

\section{METHODOLOGY}
\label{sec:methodology}
\begin{figure}[b]
    \centering
    \includegraphics[width=0.5\textwidth]{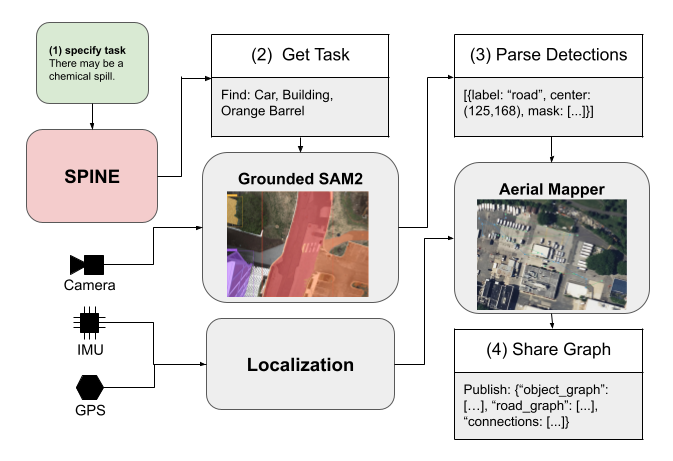}
    \caption{UAV Mapping system with dynamic input from the UGV. (1) A user defines a task which is passed to the SPINE planner. (2) The SPINE Planner assigns the UAV semantic classes to look for. (3) The UAV mapper parses the detections from Grounded SAM2 to produce a map. (4) The map is shared with the UGVs.}
    \label{fig:uav-sys}
    \vspace{-.2cm}
\end{figure}

\begin{figure*}[!ht]
    \centering
    \begin{minipage}[t]{0.3\linewidth}
        \centering
        \includegraphics[width=0.95\textwidth]{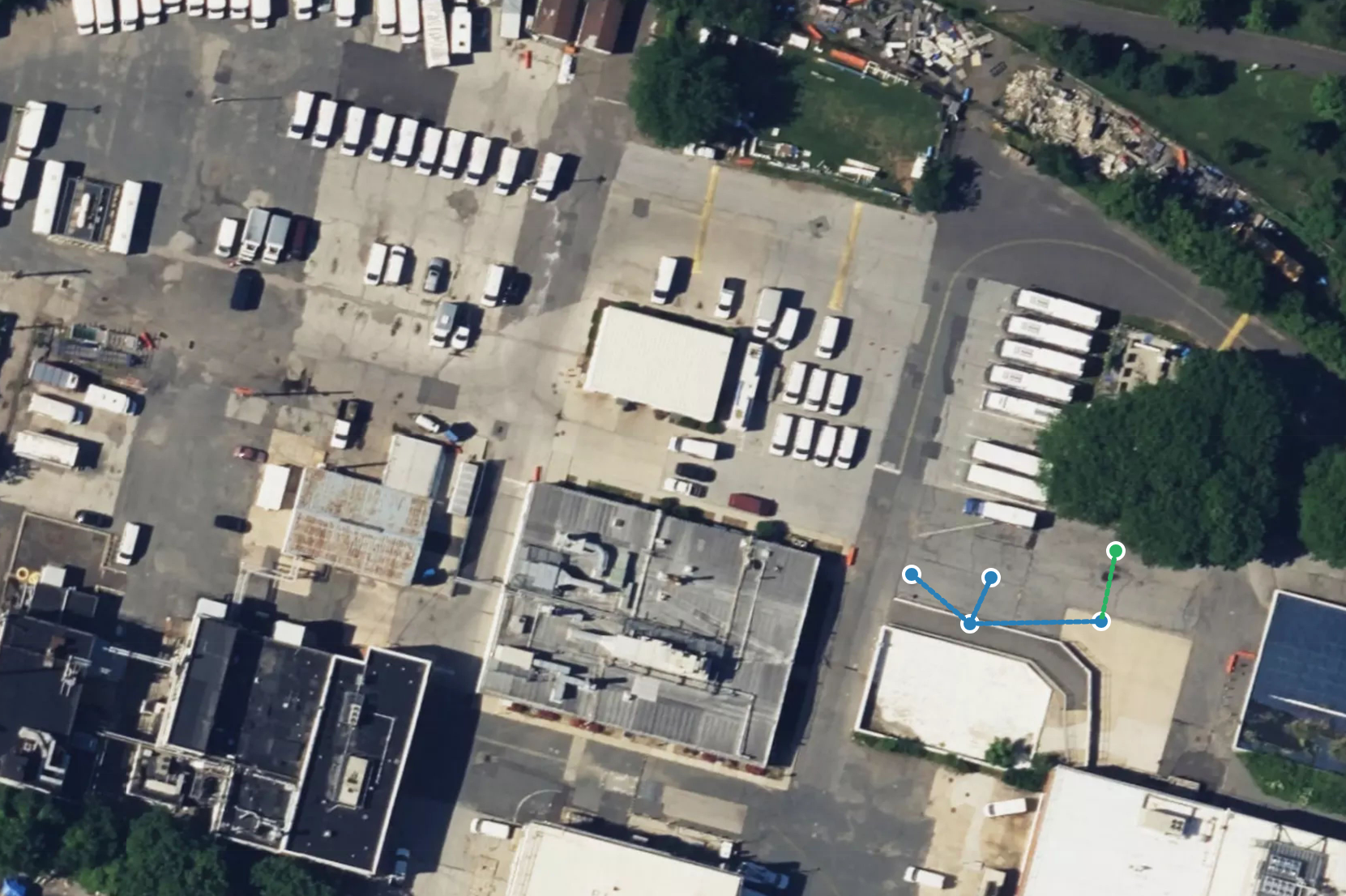}
    \end{minipage}%
    ~
    \begin{minipage}[t]{0.3\linewidth}
        \centering
        \includegraphics[width=0.95\textwidth]{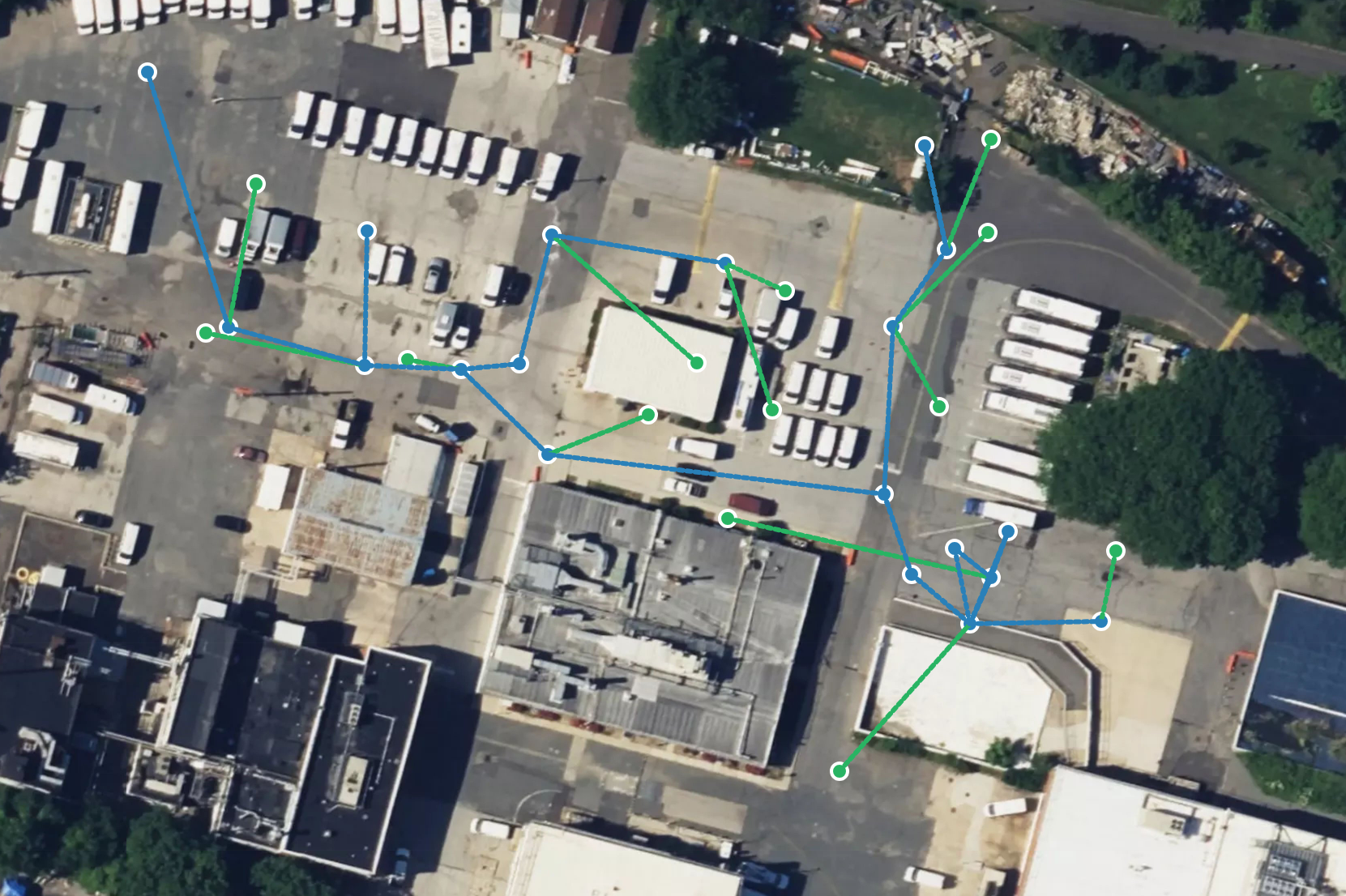}
    \end{minipage}
    ~
    \begin{minipage}[t]{0.3\linewidth}
        \centering
        \includegraphics[width=0.95\textwidth]{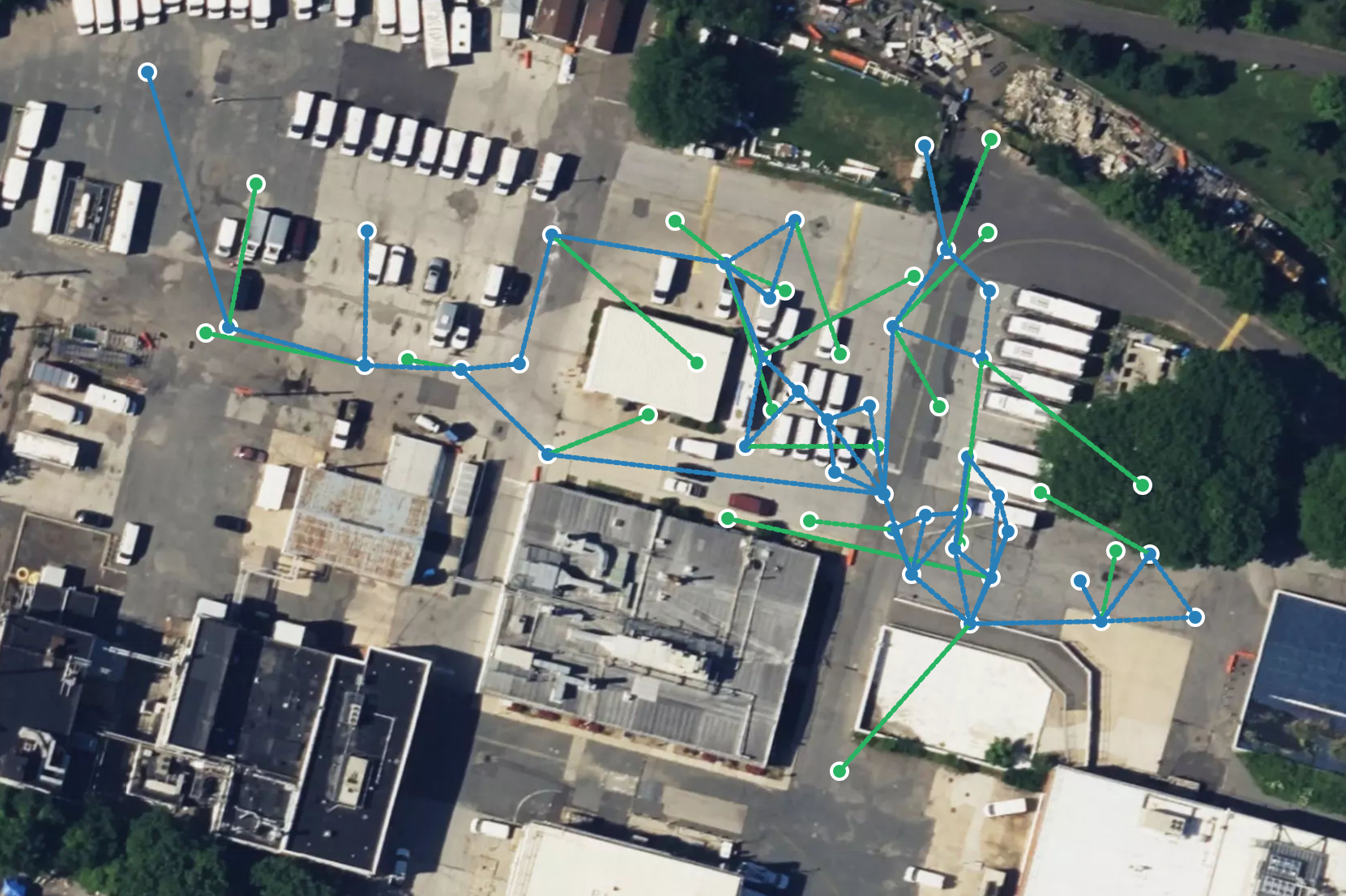}
    \end{minipage}
    \caption{An example of the generated semantic-metric map from the aerial robot is shown at different mapping iterations overlayed on Mapbox \cite{Mapbox} satellite imagery.
    The estimated traversable graph is shown in blue and the semantics (in this example \texttt{orange barrel}) is shown in green.
    \uline{Left}: Mapping iteration=4, \uline{Center}: Mapping iteration=12; \uline{Right}: Mapping iteration=25.}
    \label{fig:incremental-construction}
    \vspace{-.3cm} 
\end{figure*}

This section describes the components running onboard the \gls{uav} and \gls{ugv} and how they complement each other.
A key component of the approach is to use an \emph{open-set} semantic-metric map that can be shared between the robots (compact communication) and efficiently supply updates in a structured format to the LLM-based decision maker.
The core representation used by SPINE on the \glspl{ugv} is a semantic-topological graph.
As is common in the semantic mapping literature, each node is either a \textit{region} or an \textit{object}.
Regions are traversable points in freespace~\cite{hughes2024foundations, werby23hovsg}.
An edge between two regions indicates that there is an obstacle-free path.
Localized objects are represented as object nodes.
An edge between an object and a region node indicates that the object can be observed from that region.
Regions and objects may be enriched with additional semantic information (\eg~this bridge is blocked by a vehicle), which provides additional context for planning.
Both \gls{ugv} and \gls{uav} mapping approaches use Grounding DINO~\cite{liu2023groundingdino} to detect objects of interest and incorporate information into the map to account
for their particular visual and compute capabilities, which are described in the following sections.

\subsection{UAV Semantic Mapper}
\label{sec:bev}
This module is designed to build an object-centric semantic map that is grounded in a common frame of reference.
It also maintains a region network of the environment that the \glspl{ugv} can use for global planning.
The components of the \gls{uav} semantic mapper are shown in Fig.~\ref{fig:uav-sys}, and an example of such a map is shown in Fig.~\ref{fig:incremental-construction}.
For the purpose of this work, we assume that the traversable component for the \glspl{ugv} is the semantic class \texttt{road}. Still, our system can handle other classes if desired or required by the operation environment or mission.

At the start of each mission, the set of labels is communicated to the aerial robot using MOCHA.
The mapper then listens for images and their associated \gls{gnss} location and orientation from a custom GNSS/INS system.

\medskip \noindent \textbf{Localization. }
From our past work~\cite{miller2022stronger, cladera2024evmapper}, a major challenge in building useful aerial maps is the quality of orientation estimation and synchronization between the imagery and the odometry.
In high wind scenarios where the \gls{uav} may maintain a significant pitch, localized points in the image projected onto the ground plane may be incorrect by several meters.
We use GTSAM~\cite{dellaert2012factor} to fuse \gls{gnss} coordinates and IMU data to get a high rate state estimate between \gls{gnss} readings.
This allows us to get an accurate pose for each image.
Further details on the implementation can be found in the Appendix.~\ref{sec:app_localization}

\medskip \noindent \textbf{Object Detection and Masking.}
We use Grounded SAM2~\cite{ren2024groundedsamassemblingopenworld} for aerial object detections because of its zero-shot open-set performance, even on aerial data.
The model uses Grounding DINO~\cite{liu2023groundingdino} for open-set object detection and SAM2 \cite{ravi2024sam2segmentimages} for accurate segmentation of the objects.
We use a base and large model, respectively, which combine to 12~GB of VRAM, with a latency of 3 to 5 seconds when running onboard the \gls{uav}, depending on how many objects are detected.

\medskip
\noindent \textbf{Performance vs Compute Tradeoffs. }
Our system allows for abstract task specifications.
This underspecified nature can result in several objects of interest being identified for mapping.
We notice empirically that when smaller objects need to be detected that are harder to see from the air, the memory requirements of the detection model exceeds the available memory.
In this case, we optionally use a smaller segmentation module SlimSAM~\cite{chen20230} to ensure that the compute stays within budget.
We note that this will result in noisy locations of objects being added to the graph but the \gls{ugv} can use it's additional compute capabilities to verify and correct the graph while still ingesting a coarse prior.
The implications of this trade off is discussed in Sec.~\ref{sec:methodology}.\ref{subsec:spine_mapper} and demonstrated in Sec.~\ref{sec:sys_demo}.

\medskip \noindent \textbf{Graph Construction.}
Given the detections, the object coordinates are estimated by applying a distance transform on their associated pixel centroids.
Each object is assigned the furthest point from the mask boundary as its coordinate.
The pixel coordinates are then converted to UTM for a pre-specified origin with known camera intrinsics.
To construct  a region graph we use \texttt{road} as the semantic class. We first take the largest road mask in each image and only add a region node if the mask is more than 20\% of the image.
We also use the class \texttt{building} and \texttt{car} as negative objects, i.e. objects that we do not report. Without them the network often misclassifies buildings and cars as road, leading to untraversable edges in the graph.
These filtering methods helps reduce the number of misclassifications of other objects in the scene as roads.
When there is a large road segment we compute the same distance transform as before and assign the road point to the furthest point from the mask boundary.
This point is then transformed to UTM coordinates and added to the region graph network by connecting it to the nearest region node. This pipeline is shown in Fig. \ref{fig:uav-sys} and results of the aerial semantic graph are shown in Fig. \ref{fig:incremental-construction}.

\subsection{UGV Semantic Mapper}
\label{subsec:spine_mapper}

The \gls{ugv} mapper also maintains a local occupancy map, which the planner uses for free space exploration.
The semantic graph constructed by the \gls{uav} provides an initial map estimate (see Sec. IV~\ref{sec:bev}), and the \gls{uav} can iteratively provide map updates during the mission.

Our \gls{ugv} semantic mapping implementation is shown in Fig.~\ref{fig:sys-overview}.
The mapper takes RGB + Depth, LiDAR, and semantic configuration as inputs.
LiDAR is used for odometry estimation (Faster-LIO~\cite{fasterlio}) and local occupancy map construction (GroundGrid~\cite{GroundGrid}).
The occupancy map is used to add and remove regions and edges from the map based on connectivity.
RGB+D is used for object localization and captioning.
Objects are detected using GroundingDino~\cite{liu2023groundingdino}).
Detections are then clustered and localized with a multiple-hypothesis tracker.
A vision-language model (LLaVA~\cite{liu2023llava}) enriches the semantic information available to the planner (see  Fig.~\ref{fig:buckner_mission}).
Outputs from these modules are used to add and remove nodes and enrich them with semantic information.
A Semantic configuration is provided by the planner and is used to set the labels of the object detector and provide queries to the vision language model.
The detection and tracking modules run at $\sim 5\,\text{Hz}$, the vision-language model runs at $\sim 1\,\text{Hz}$, and occupancy map construction runs well over $10\,\text{Hz}$, all onboard.

\subsection{SPINE Planning}
\label{sec:spine_llm_planner}

The backbone planner used in this work is based on SPINE, first presented in ~\cite{ravichandran_jackal}.
This section summarizes some of the key ideas relevant to teaming.

SPINE's plan generator uses an LLM to infer a task sequence from a mission specification and semantic map.
We configure a pre-trained LLM via a system prompt with three components: role description, the mapping interface, and a description of the behavior library.

\medskip \noindent \textbf{Mapping Interface.}
The mapping interface provides a textual representation of the semantic graph.
The plan generator first receives a JSON representation of the graph. Then, at
each planning iteration, all map updates are provided to the LLM in-context via the following API, which captures high-level graph manipulations: \verb|add_nodes|, \verb|remove_nodes|, \verb|add_edges|, \verb|remove_edges|, and \verb|update_nodes|.
The nodes are defined as a dictionary of attributes, which allows for providing nodes with rich semantic descriptions (example in Fig.~\ref{fig:vlm_example}).

\medskip \noindent \textbf{Plan Generation.}
SPINE composes plans via behaviors for navigation, active mapping, and user interaction.
The LLM generates a receding horizon behavior sequence at each planning iteration, and this sequence is provided to the validation module.
An ``answer'' behavior terminates a mission and notifies the user of results, and
a ``clarify'' behavior is used to gain further instructions, if needed.

\begin{figure}[t]
    \centering
    \includegraphics[width=0.95\linewidth]{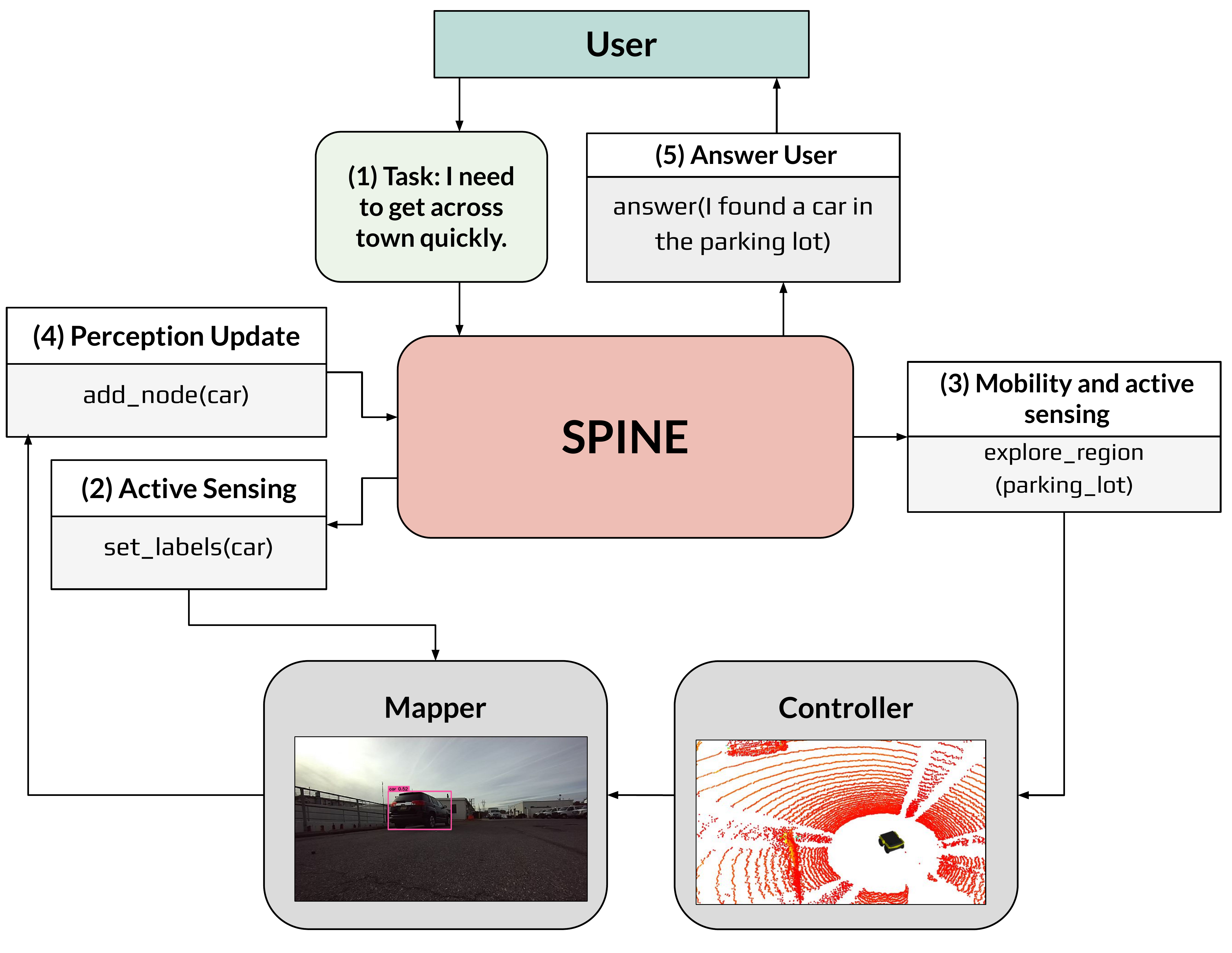}
    \caption{Example API use during mission. (1) user specifies a mission. (2) SPINE infers relevant semantic categories and configures perception labels. (3) SPINE then reasons about exploration targets. (4) Perception finds a car and notifies planner. (5) Planner informs user.
    }
    \label{fig:action-perception-loop}
    \vspace{-.2cm}
\end{figure}

We enforce chain-of-though reasoning by requiring the LLM to provide a justification for each action sequence, which reduces hallucinations or otherwise ungrounded behavior~\cite{cot_llm}.
All inputs to the planner and action history are maintained in-context via the provided APIs.
See Fig.~\ref{fig:action-perception-loop} for an example action-control-perception loop.

\medskip

\noindent \textbf{Plan Validation.}
To create subtasks, the planner must correctly invoke its behavior library while reasoning over constraints such as traversability.
Since LLMs are prone to hallucinate this information, we filter LLM-generated plans through a validation module,
 which ensures all plans are syntactically correct and physically realizable.
If a given task is invalid, the validator forms state-specific feedback to the LLM.
While this validation procedure provides
valuable planning safeguards, the spatial constraints are contingent on perception.

\medskip

\noindent \textbf{Online Map Correction.}
SPINE uses prior semantic graphs constructed from a \gls{uav} or satellite image.
Such priors offer valuable contextual and traversability information, however they may contain spurious traversability edges.
Identifying such errors online so that SPINE can replan is vital to a robust autonomy solution.
SPINE thus uses feedback from the downstream controllers to infer spurious edges and remove them from the graph.
If SPINE sends a navigation command across an edge that the controller cannot traverse, SPINE removes that edge from the graph.

\begin{figure}[!t]
    \centering
    \includegraphics[width=0.85\linewidth]{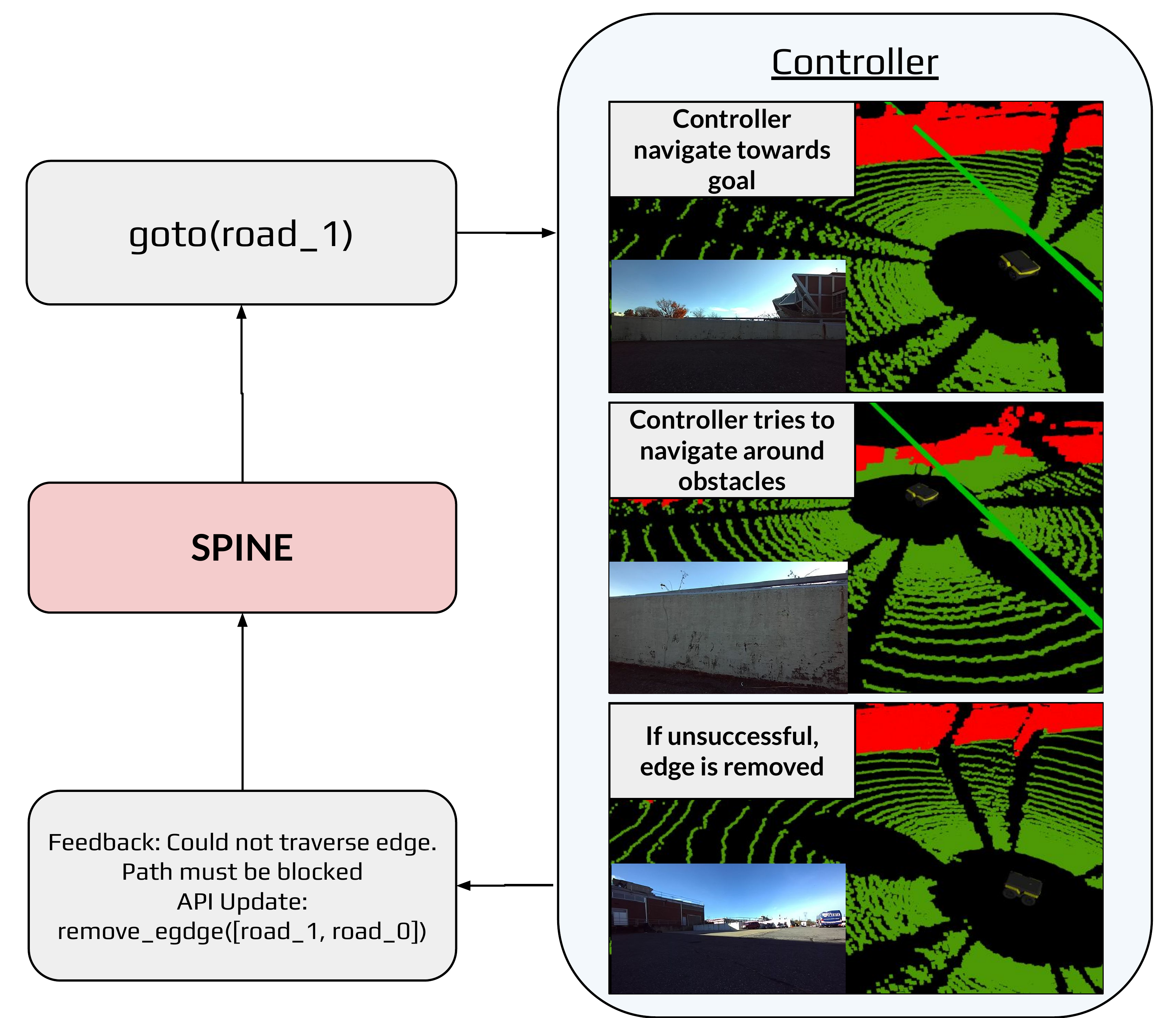}
    \caption{Example map correction. SPINE attempts to traverse an invalid edge (green line). Green points denote ground and red points denote obstacles. The controller cannot find a trajectory around the obstacle, and it times out after a pre-specified duration. The incorrect edge is removed from the graph, and that feedback is sent to SPINE.}
    \label{fig:spine_online_map_correction}
    \vspace{-.2cm}
\end{figure}

\subsection{Mission Execution}
The \gls{uav} is tasked with an initial exploration mission over the area of interest.
For our experiments, a pre-defined waypoint set is provided to avoid flying over unsafe regions in the area.
However, waypoints can be generated with any coverage algorithm.

After a timer $t_i$ has elapsed, the \gls{uav} transitions into search and communication modes, looking for ground robots to transmit the graph.
The last known pose is used as a heuristic for where to find the ground robot.
Once the ground robot has been found and all the data has been transmitted, the \gls{uav} transitions again into exploration mode, covering all the waypoints in its mission.

On the \gls{ugv}, SPINE sends navigation subtasks to the low-level planner.
These commands reference paths on the semantic graph (note that exploration commands first extend the semantic graph).
Graph navigation paths are then sent to the ROS Move Base controller~\footnote{\url{http://wiki.ros.org/move_base}}, which plans trajectories realized by the Jackal controller.

\section{EXPERIMENTS}
\label{sec:experiments}
\begin{figure}[!t]
    \centering
    \includegraphics[width=0.95\linewidth]{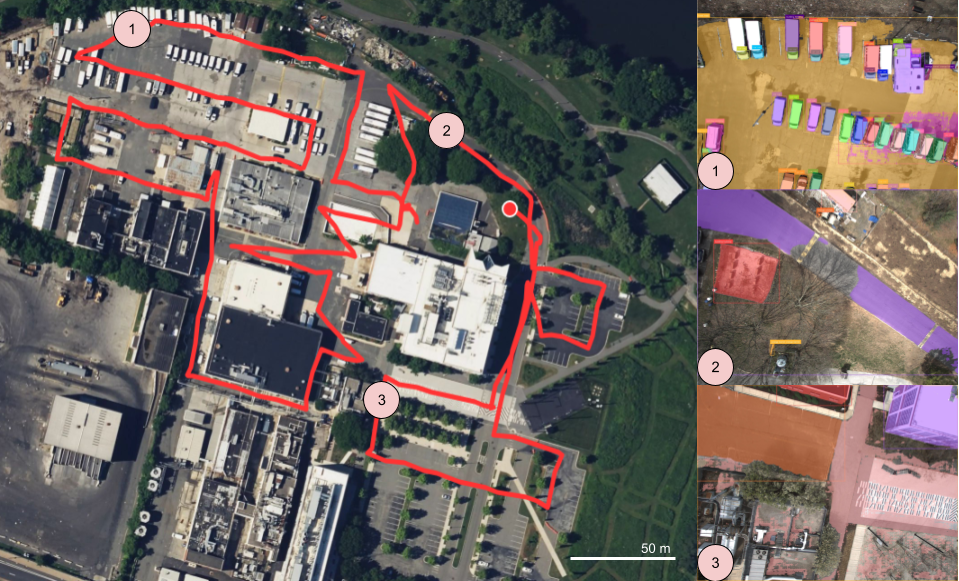}
    \caption{Example images of the detection onboard the aerial robot for the Pennovation experiment, showing the quality of the zero-shot transfer to the aerial data. The red line indicate the flight path of the UAV overlayed on Mapbox \cite{Mapbox} satellite imagery.}
    \label{fig:qualitative-bev}
    \vspace{-.2cm}
\end{figure}

We evaluate our system over three sets of experiments.
We first evaluate the incremental graph construction of the Aerial Autonomy component of our system (Sec. V~\ref{sec:air_experiment}).
We then evaluate the Ground Autonomy's ability to realize language-specified missions in partially-known environments (Sec. \ref{sec:experiments}.\ref{sec:ground_experiment}).
Finally, we demonstrate air-ground teaming in unknown environments given language-specified missions, in controlled settings  (Sec \ref{sec:experiments}.\ref{subsec:air-ground-evaluation}).

\medskip

\noindent \textbf{Platforms. }
We used the Falcon 4~\cite{canopy} as our \gls{uav} platform.
The UAV is controlled by a
flight controller running PX4 firmware~\cite{meier2015px4}.
The primary sensors are a Blackfly S U3-32S4C-C RGB global-shutter camera, and a VectorNav VN-100-T \gls{imu}.
The platform uses a U-blox ZED F9P \gls{gnss} sensor for localization.
Finally, all the data is processed by an onboard Nvidia Jetson Orin NX.

Our \gls{ugv} platform consists of a Clearpath Jackal equipped with an AMD Ryzen 3600 CPU with 32 GB of RAM and an Nvidia RTX 4000 Ada SFF GPU.
An Ouster OS1-64 LiDAR is used for obstacle avoidance, path planning, and odometry, while a ZED 2i stereo camera provides sensing for object detection.
We also use a U-blox ZED-F9P \gls{gnss} to record the initial position of the \gls{ugv}, which establishes a common reference frame with the overhead graph.
Finally, we use a Vectornav VN-100 for global heading corrections.
More details about the hardware platforms can be found in the Appendix.

\medskip

\noindent \textbf{Environments.} We perform Ground Autonomy experiments in two rural environments termed Bucker and Range 15, shown in Fig.~\ref{fig:buckner_mission} and Fig.~\ref{fig:r15_mission}, respectively.
Each environment stresses a unique component of the autonomy stack.
Range 15 is large and semantically sparse.
We use this environment to evaluate the autonomy's reasoning over long-horizon missions, state estimation, and communications at scale.
Buckner is smaller in scale but features richer semantics, such as bridges, gates, and parking lots, all within a few hundred meters.
Buckner experiments evaluate the UGV autonomy's ability to reason over richer or
ambiguous specifications.

We demonstrate the full air-ground system in Pennovation, an urban office park with fields, buildings, parking lots, and other structures.
We also use data from these demonstrations to evaluate the aerial autonomy performance.
All environments were uncontrolled.
There were dynamic entities such as cars and people moving during experiments,
and the environments contained many obstacles, both positive (fences, walls) and negative (ditches).

\subsection{Aerial Mapping}
\label{sec:air_experiment}
\begin{figure}[b]
    \centering
    \includegraphics[width=0.95\linewidth]{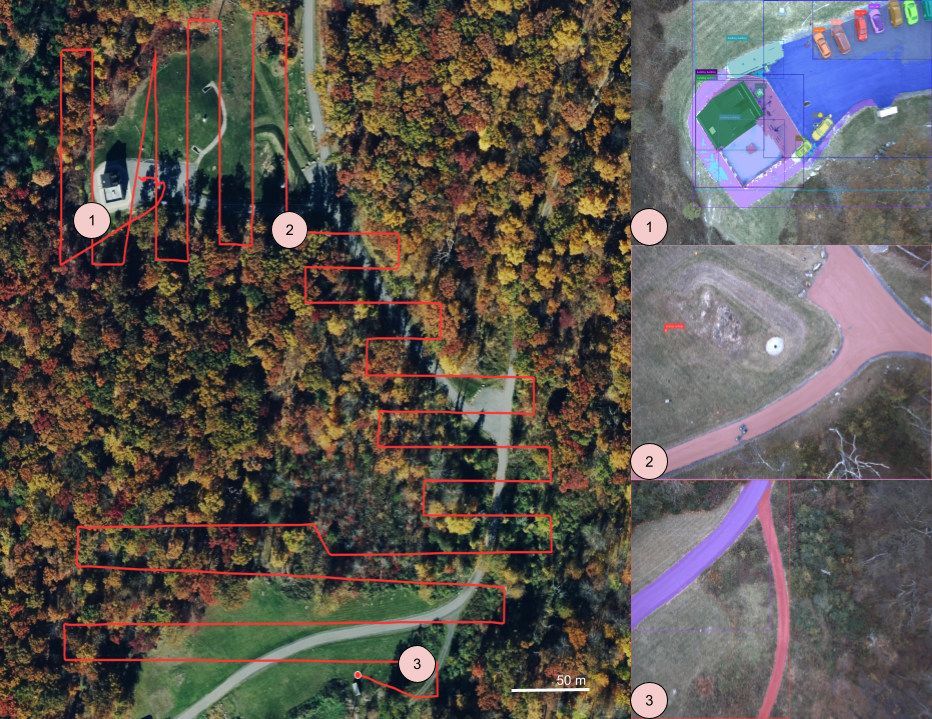}
    \caption{Semantic detection results for Range 15, showing the zero-shot transfer capabilities of Grounded-SAM2 to a rural environment. The read line shows the flight path of the UAV overlayed on Mapbox \cite{Mapbox} satellite imagery.}
    \label{fig:qual_range_15}
\end{figure}

In this section, we evaluate the components of the \gls{uav} mapping system.
The important qualities of our aerial maps are:
\begin{enumerate*}
    \item Resolution, task relevance, semantic detections;
    \item Small size for transmission between robots;
    \item Accurate estimate of the traversable portions of the environment.
\end{enumerate*}
We design experiments to measure the effectiveness of our proposed method for these qualities.
We first test the quality of the object detector, and we then measure the size of the stored maps relative to the distance traveled.

\medskip

\noindent \textbf{Object Detections.}
To evaluate our object detection module, we manually annotate the desired objects in 2 of our datasets and evaluate the false positive rate of the detector.
The desired labels for this evaluation are task specific to these datasets and are not the same across both datasets.
The results of this evaluation are shown in Tab.~\ref{tab:object_detections}.
As shown, the object detector is able to perform remarkably well with zero-shot transfer to our data.
The results are surprising considering the viewpoint of the \gls{uav} camera.
The qualitative evaluation of the detected objects is shown in Fig.~\ref{fig:qualitative-bev} \& Fig.~\ref{fig:qual_range_15}.

\begin{table}[t]
    \begin{center}
    \begin{tabular}{cccc} \toprule
         Specification & Total Detections & False Positives & Precision \\ \toprule
         Pennovation 1  & 708 & 127 & 82.1 \% \\
         Pennovation 2 & 388 & 59 & 84.8 \%\\
        \toprule
    \end{tabular}
    \caption{Performance of the aerial object detection module based on Grounded SAM2, zero-shot transfer.}
    \label{tab:object_detections}
    \end{center}
    \vspace{-.8cm} 
\end{table}

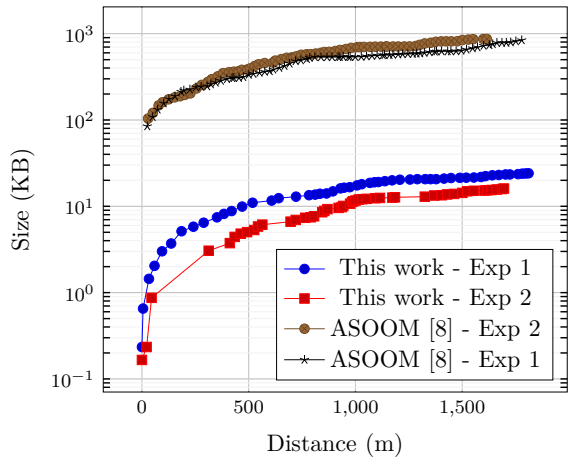
\begin{figure}[!b]
        \begin{tikzpicture}[scale=0.9]
    \pgfmathsetlengthmacro\MajorTickLength{
      \pgfkeysvalueof{/pgfplots/major tick length} * 0.05
    }
    \pgfplotsset{every tick label/.append style={font=\footnotesize}}
    \begin{axis}[xlabel={Distance (m)},
                 ylabel={Size (KB)},
                             ymode=log,
                 major tick length=\MajorTickLength,
                 grid=both,
                 grid style={line width=.1pt, draw=gray!10},
                 major grid style={line width=.2pt,draw=gray!50},
                 legend pos=south east,
                 every tick/.style={
                    black,
                    semithick,
      }]

    \addplot table[x index=0,y index=1,col sep=comma] {data-pennov1.dat};
    \addlegendentry{This work - Exp 1}

    \addplot table[x index=0,y index=1,col sep=comma] {data-pennov2.dat};
    \addlegendentry{This work - Exp 2}

    \addplot table[x index=0,y index=1,col sep=comma] {3robots5_combined.dat};
    \addlegendentry{ASOOM~\cite{miller2022stronger} - Exp 2}

    \addplot table[x index=0,y index=1,col sep=comma] {3robots9_combined.dat};
    \addlegendentry{ASOOM~\cite{miller2022stronger} - Exp 1}
    \end{axis}
    \end{tikzpicture}
    \caption{Size of the uncompressed map as a function of the distance traveled by the \gls{uav}.}
    \label{fig:uav-graph-size-vs-distance}
\end{figure}

\noindent\textbf{Incremental Graph Construction.}
To evaluate the quality of the constructed graph, we show an example of the incremental semantic-metric map generated by the robot overlaid onto a satellite image in Fig.~\ref{fig:incremental-construction}.
An advantage of our method is the size of the map that is stored.
We show in Fig.~\ref{fig:uav-graph-size-vs-distance} that the maps are roughly linear in the distance traveled, and areas of size 20,000 m$^2$ can be stored in just kilobytes.

Compared to our previous work~\cite{miller2022stronger, spomp_journal},  the messages transmitted between robots are significantly smaller. For instance, the size of the uncompressed graph after flying 1000 m is approximately 12 to 17 KB, compared to 530 to 699 KB of the semantic map image generated by ASOOM.
These results showcase the efficiency of the aerial semantic graph generated by the \gls{uav} compared to the previously used dense maps.

\medskip

\noindent\textbf{Traversability Estimation.}
Finally, to measure the quality of the estimated road network, we manually annotate the true positives and false positives in the detected roads.
The results of this analysis are shown in Table~\ref{tab:traversal-estimation}. These results are encouraging: accurate road maps improve the performance of the ground autonomy stack.

\begin{table}[t]
    \begin{center}
    \begin{tabular}{cccc} \toprule
         Specification   & Total Detections & False Positives & Precision \\ \toprule
         Pennovation 1 & 82 & 9 & 89.0 \% \\
         Pennovation 2 & 24 & 2 & 91.6 \%\\
        \toprule
    \end{tabular}
    \caption{Performance of the traversable edge detection module. }
    \label{tab:traversal-estimation}
    \end{center}
    \vspace{-.8cm} 
\end{table}

\subsection{Ground Autonomy}
\label{sec:ground_experiment}
We evaluate the UGV autonomy platform independently of the UAV.
We design experiments to assess the UGV autonomy's ability to complete missions with differing specifications, environments, and priors.
In contrast to the System Demonstration in Sec. V~\ref{subsec:air-ground-evaluation}
and~\ref{sec:sys_demo}, we generate the priors from registered satellite data.
The priors contain traversability information and some relevant semantics.
However, the priors are imperfect and contain irrelevant information and mistakes that the UGV must correct online.
Our experiments evaluate the UGV's ability to infer navigation, exploration, and information acquisition goals from natural language specifications, react to findings, and use findings to complete the user's mission.

\medskip
\noindent\textbf{Experimental Setup.} We consider five mission specifications, as summarized in Table~\ref{tab:ugv_specs}, and evaluate each specification one to four times.
We construct priors for each environment; however, the priors are not designed for a specific tasks.
One of the primary purposes of the prior was to keep the robot away from negative obstacles, such as ditches, which could not be detected by the \gls{ugv}.

\begin{table*}[t]
    \begin{tabular}{p{4.5cm}cp{4cm}p{5.5cm}} \toprule
         Specification & Location & Priors & Desired outcome\\ \toprule
         (S1) I heard of activity near the red houses. Go check & Range 15  & Red houses. No objects & Identity and report several people near two red houses. \\
         (S2) Go inspect the black vehicle at the end of the driveway  & Range 15 &  Black vehicle is at the beginning of the driveway & Planner identifies misspecification and correctly finds car.  \\
         (S3) I got reports that the bridge was blocked. Go check. & Buckner & Bridge, no blockage & Planner identities vehicle blocking the bridge \\
         (S4) Is there activity near the gate & Buckner &  Gate, objects & Planner identifies and reports a person \\
         (S5) Identify the parking lot 30 meters east of the bridge & Buckner & Bridge, but no parking lot & Planner explore correct area and identifies parking lot \\
        \toprule
    \end{tabular}
    \caption{Mission specifications, locations and priors used to test the ground autonomy.}
    \label{tab:ugv_specs}
    \vspace{-.3cm} 
\end{table*}

\medskip
\noindent\textbf{Results and Discussion.}
We summarize the experimental results for each mission specification in Tab~\ref{tab:ugv_outcomes}.
We report mission success, distance traveled, and the number of API calls.
The number of API calls indicates how many subtask steps SPINE inferred during its mission.

The UGV was successful in achieving S2 through S5.
There were two unsuccessful missions during S1, each of which failed due to the large scale of the environment.
In one outcome, a gust of leaves blew around the ground vehicle's LiDAR, which caused unrecoverable odometry drift (see Fig.~\ref{fig:ugv_odom_fail}).
In the second, the UGV lost communications and could not perform the LLM queries required for planning.

\begin{table}[b]
    \begin{tabular}{ccccc} \toprule
         Spec.   & Outcomes & Dist. (m) & API Calls & Failure modes\\ \toprule
         S1 & 1/3 & 1200 & 2 & Odom., Comm. \\
         S2 & 2/2 & 231  & 4 & N/A \\
         S3 & 4/4 & 265 & 2 &  N/A \\
         S4 & 1/1 & 132  & 2 &  N/A \\
         S5 & 1/1 & 450 & 3 &   N/A \\
        \toprule
    \end{tabular}
    \caption{UGV autonomy outcomes. SPINE completed all missions except for two during the first specification, which failed due to odometry drift and communication loss.}
    \label{tab:ugv_outcomes}
\end{table}

\begin{figure}[t]
    \centering
    \includegraphics[width=0.99\linewidth]{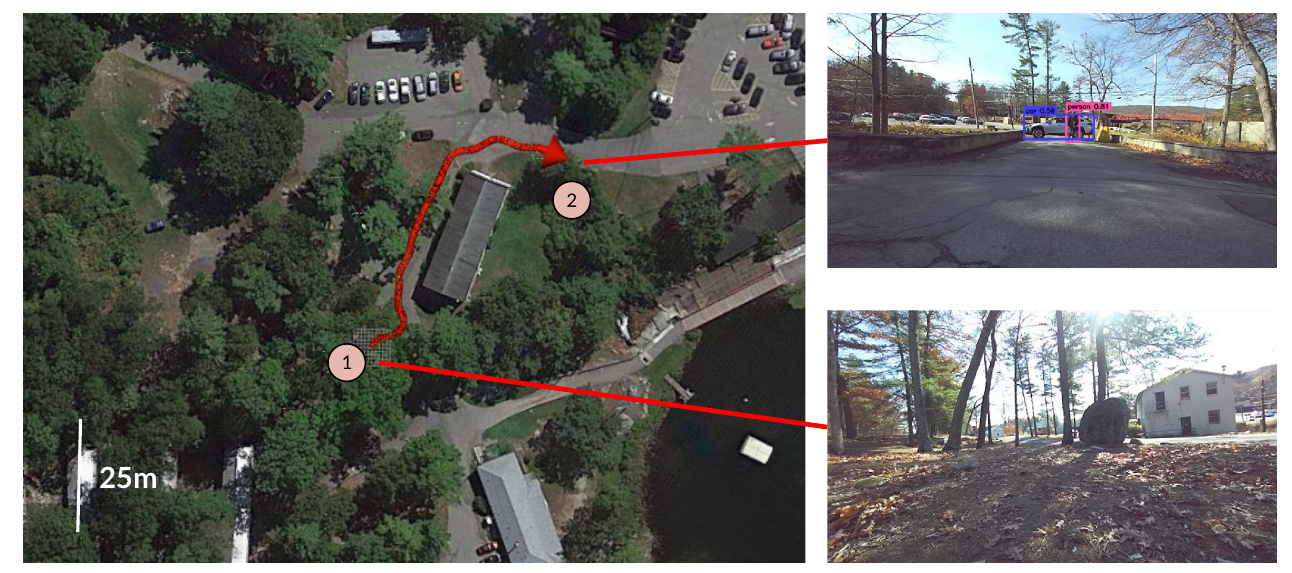}
    \caption{Example mission from S3. ``I got reports that the bridge was blocked. Go check.'' The \gls{ugv} starts its mission at the bottom left of the trajectory (1). The \gls{ugv} then infers an mapping target (bridge). After navigation and mapping, it identifies several obstacles and reports its findings to the user (2).}
    \label{fig:buckner_mission}
\end{figure}

\begin{figure}[b]
    \centering
    \includegraphics[width=0.8\linewidth]{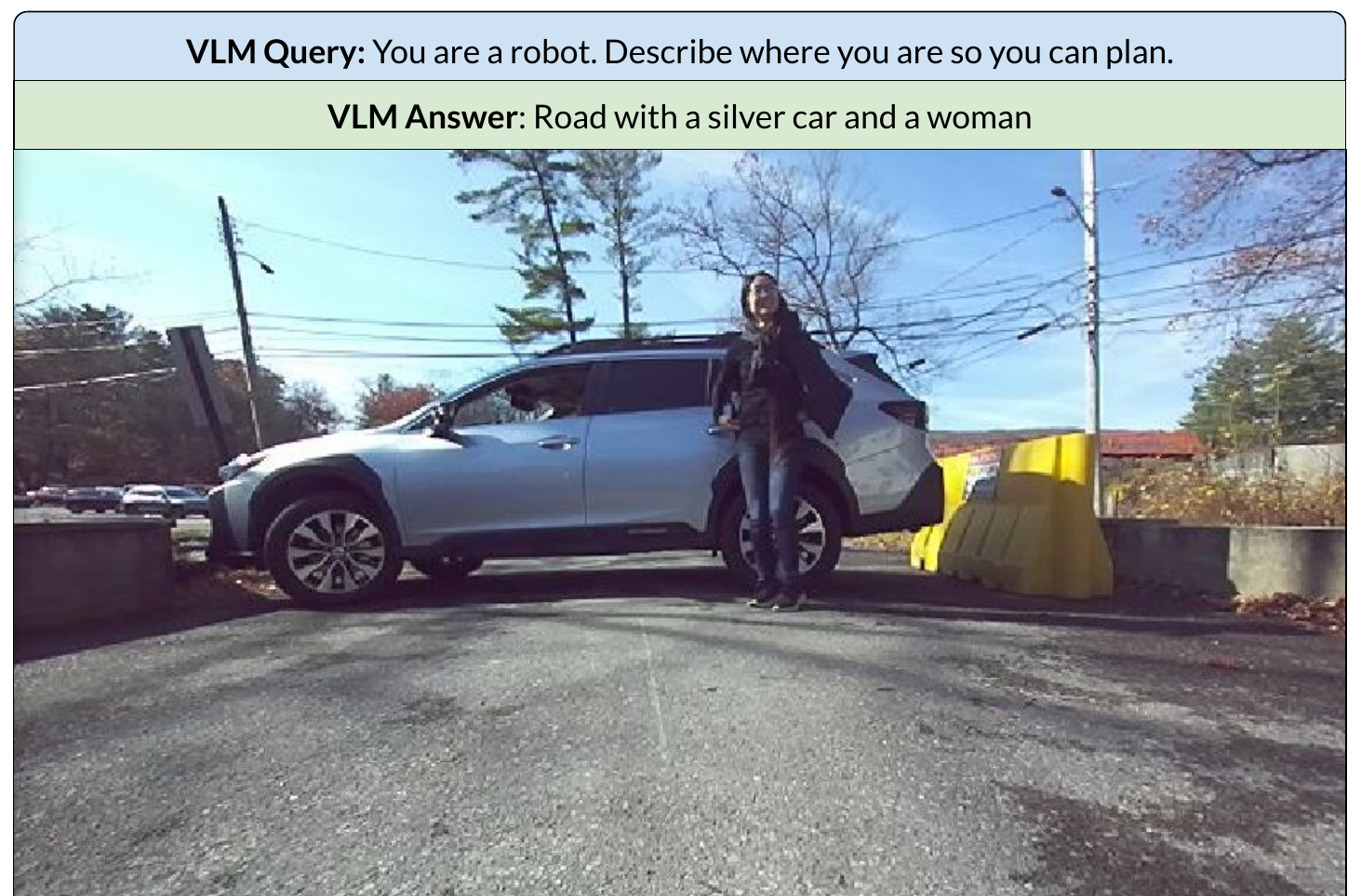}
    \caption{In the above example, the ground vehicle must learn if the bridge is blocked. The ground vehicle (Specification 3). The ground vehicle's planner, SPINE, uses a Vision Language Model running onboard to obtain detailed semantic information for the user. }
    \label{fig:vlm_example}
\end{figure}

We highlight two emblematic missions.
Fig.~\ref{fig:buckner_mission} shows the UGV trajectory from a mission with the specification: ``I heard the bridge was blocked. Can you check?''.
SPINE infers that the relevant semantics are bridge, car, truck, bicycle, pedestrian, construction cone, debris, and barrier.
The planner then navigates to the bridge by constructing the following plan \verb|goto(road_5), map_region(bridge_1)|.
Upon mapping the bridge, SPINE discovers several obstacles, including a person, car, and construction barrier.
SPINE uses the VLM to get a more detailed description, as shown in Fig.~\ref{fig:vlm_example}, and reasons over this information to respond ``The path across the bridge is likely blocked due to the presence of a construction cone and a person. There is also a silver car and a woman on the bridge, which may indicate a temporary blockage.''

\begin{figure}[t]
    \centering
    \includegraphics[width=0.9\linewidth]{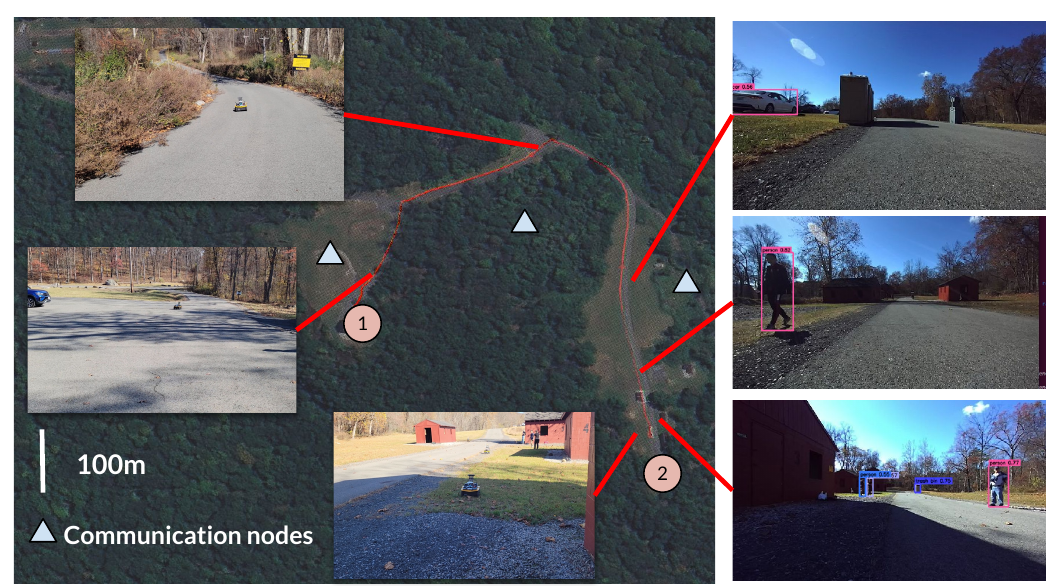}
    \caption{Example mission from S1. ``I heard of activity near the red houses. Go check.'' At the start (1), the \gls{ugv} identifies exploration targets (red houses, bottom) and navigates there from its start location (2). The \gls{ugv} successfully identifies several people. The ground vehicle communicates its findings in realtime via text. It then returns to its starting location to offload mission data.}
    \label{fig:r15_mission}
\end{figure}

Fig.~\ref{fig:r15_mission} overviews a mission in which the \gls{ugv} was provided the task: ``I heard of suspicious activity near the red houses. Go check.''
The \gls{ugv} had red houses on its prior map, so it infers a plan to map those.
Along the way, it observes cars and people. Because those objects are not near the red houses, the UGV records those in its map but does not consider them as relevant to the task.
Finally, the UGV reaches the red houses, discovers people, and reports them to the user.
The user then asks the UGV to return to the starting location to offload data. Overall the mission requires the UGV to traverse 1200 meters, which test all layers of the autonomy stack including the odometry and communication network. SPINE uses active perception to resolve errors in the mission specification. Examples about active perception are presented in the Appendix.

\subsection{Air-Ground Teaming Evaluation}
\label{subsec:air-ground-evaluation}
We evaluate the performance of our air-ground teaming system as compared to single-UGV variants against language-specified missions in unknown environments.
The first baseline (``UGV w/ GT'') receives a full map constructed from satellite imagery, which estimates upper-bound performance of the mission given our UGV autonomy stack.
The second baseline (``UGV'') receives no prior map, and this baseline must explore to accomplish the mission.
We consider a triaging mission where the system must respond to a chemical spill by inferring relevant semantic (\textit{i.e.,} chemical barrels, people), map the relevant area, and report its findings.
The system is given the specification ``triage the chemical spill 100 meters X,'' where X is a location prior. Note the system is not explicitly provided the relevant semantics (\textit{i.e.,} barrels); rather, it must infer. We repeat this experiment three times for each baseline and six times for our system, with location priors to the north and northeast.

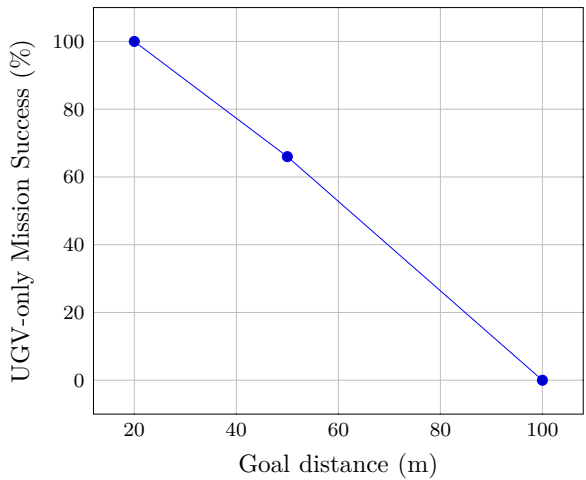
\begin{figure}[b]
    \begin{tikzpicture}[scale=0.95]
    \pgfmathsetlengthmacro\MajorTickLength{
      \pgfkeysvalueof{/pgfplots/major tick length} * 0.05
    }
    \pgfplotsset{every tick label/.append style={font=\footnotesize}}
    \begin{axis}[xlabel={Goal distance (m)},
                 ylabel={UGV-only Mission Success (\%)},
                 major tick length=\MajorTickLength,
                grid=both,
                grid style={line width=.1pt, draw=gray!10},
                major grid style={line width=.2pt,draw=gray!50},
                 every tick/.style={
                    black,
                    semithick,
      }]

    \addplot table[x index=0,y index=1,col sep=comma] {ugv.dat};

    \end{axis}
    \end{tikzpicture}
    \caption{UGV-only mission success as goal distance increases. Without priors, the UGV must build a map entirely from exploration, which it struggles to do as the goal distance increases.}
    \label{fig:ugv-exploration}
\end{figure}

We report results in Table~\ref{tab:full_experiments}, as measured by success rate, distance traveled by the UGV, and the percentage of time spent in autonomous mode.
Unsurprisingly, the UGV obtains a 100\% success rate when given a full prior map of the environments.
Our system achieves an 83.3\% success rate while spending a compatible amount of time in autonomous mode.
Our system also travels a similar  distance, indicating that the prior map provided by the UAV offers efficient paths.
Without a prior map, the UGV is unable to find the goal.
We find that the UGV increasingly struggles with long-scale exploration, as indicated in Fig.~\ref{fig:ugv-exploration}.

\begin{table}[t]
    \begin{tabular}{cccc} \toprule
         Method   & Success (\%) &  Dist. (m) & Autonomous (\%) \\ \midrule
         UGV & 0 & 54.9  & 94.6  \\
         UAV-UGV & 83.3 & 104.9 &  99.3 \\
         UGV w/ GT & 100  & 100.7 & 99.7  \\
        \toprule
    \end{tabular}
    \caption{Experiments with air-ground teaming compared to UGV-only methods. GT corresponds to a pre-computed map from satellite imagery.}
    \label{tab:full_experiments}
    \vspace{-.5cm} 
\end{table}

\section{System Demonstration}
\label{sec:sys_demo}
\begin{figure}[t]
    \begin{tikzpicture}[scale=0.95]
    \pgfmathsetlengthmacro\MajorTickLength{
      \pgfkeysvalueof{/pgfplots/major tick length} * 0.05
    }
    \pgfplotsset{every tick label/.append style={font=\footnotesize}}
    \begin{axis}[xlabel={Time (s)},
                 ylabel={Number of Nodes},
                 major tick length=\MajorTickLength,
                grid=both,
                grid style={line width=.1pt, draw=gray!10},
                major grid style={line width=.2pt,draw=gray!50},
                 every tick/.style={
                    black,
                    semithick,
      }]

    \addplot table[x index=0,y index=1,col sep=comma] {data-mission-4.dat};
    \addlegendentry{Mission 1}

    \addplot table[x index=0,y index=1,col sep=comma]  {data-mission-5.dat};
    \addlegendentry{Mission 2}

    \end{axis}
    \end{tikzpicture}
    \caption{Size of graph as a function of the mission time for the ground vehicles. Ground vehicles receive graph updates from onboard mapping and the aerial vehicle, thus the map grows substantially over the mission.}
    \label{fig:ugv-graph-size-vs-time_actual}
\end{figure}
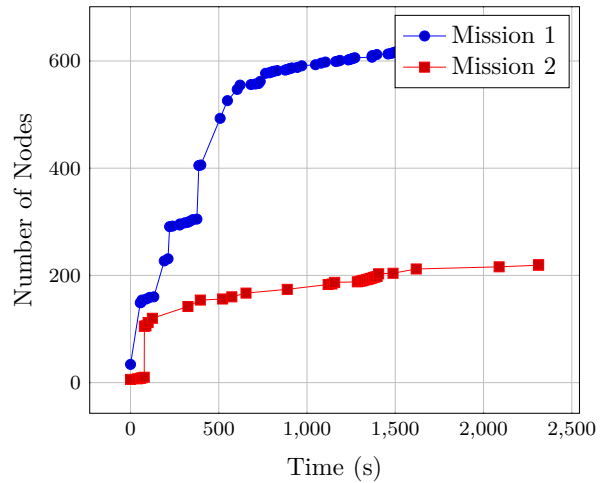
\begin{figure*}
    \centering
    \includegraphics[width=0.95\linewidth]{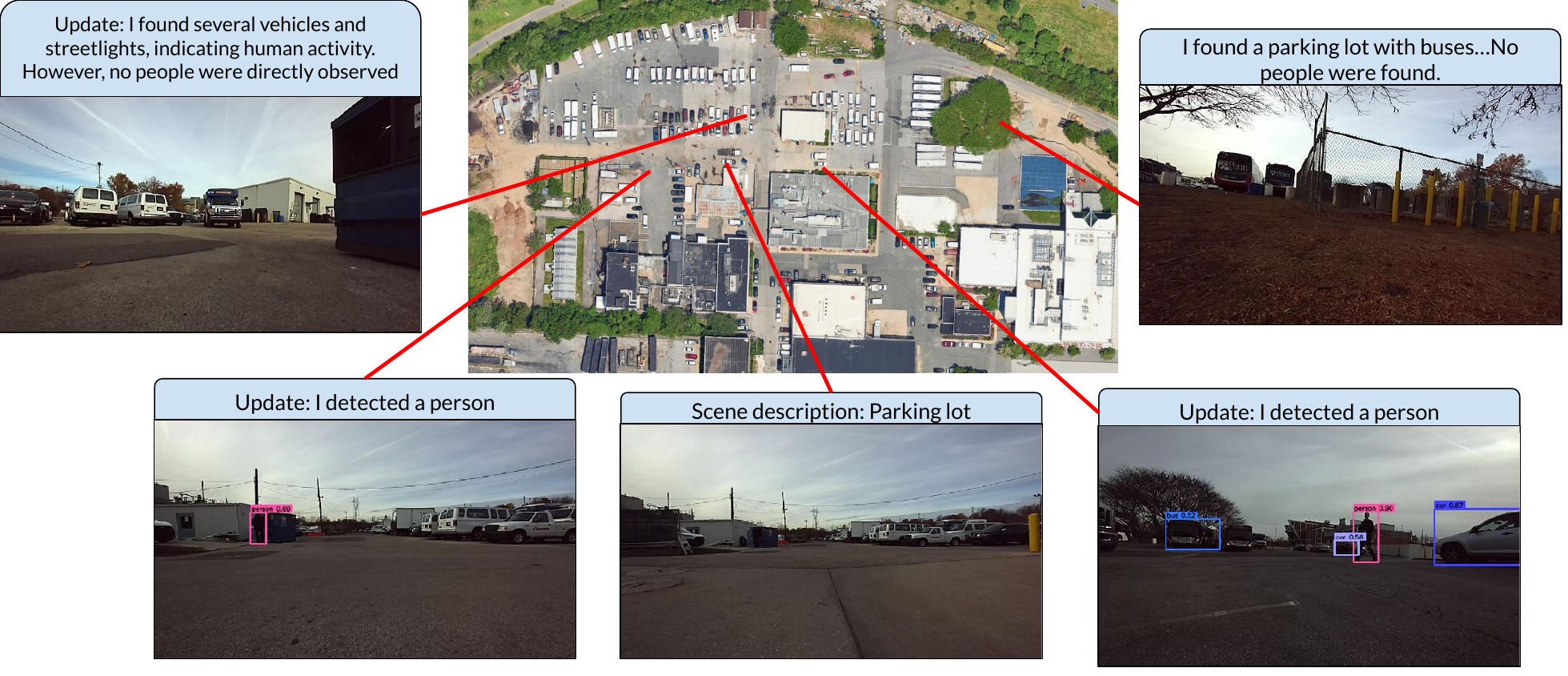}
    \caption{Snapshots of the second system demonstration. The air ground team is tasked with finding people in the Pennovation environment. The ground vehicle provides updates in natural language about mission findings (people) and broader context (scene description, etc). Because the environment is unknown at runtime, such context provides valuable situational awareness to the user.}
    \label{fig:ugv_mission_5}
\end{figure*}

Through the previous sections, we demonstrated the use of the system in relatively controlled settings.
In this section, we show qualitative demonstrations of the system in underspecified missions that require reasoning and the sub-tasks have to be inferred by the system.
As mentioned in Sec.~\ref{sec:methodology}.\ref{sec:bev}, a smaller model is used to generate the object masks to keep the compute feasible.
We demonstrate the full system through two experiments in the Pennovation environment shown in Fig.~\ref{fig:ugv_mission_5}.
The mission concept follows from the one described in Fig.~\ref{fig:mission_concept}.
Large-scale demonstrations stress the autonomy stack of the ground robots, the opportunistic communication, and the behavior of the air-ground team.
Note that in addition to the components evaluated previously, we stress the human-in-the-loop aspect of our system.
In particular, the operator is constantly receiving updates and retasking the robot.

We report the total distance traveled, mission duration, human interactions, API calls, prior edges removed, and percent of the time the \gls{ugv} was in autonomous mode.
As the system experiments feature human in the loop tasking, we report the number of user interactions.
We also report the number of API calls, the number of edges removed by the planner, and percent of time the \gls{ugv} was in autonomous mode.
All results are reported in Table~\ref{tab:sys_outcomes}.

\medskip
\noindent\textbf{System Demonstration 1: Mapping and Inspection}
The first mission was specified by the query ``I heard of construction  around the eastern roads. Can you check?.''
The planner inferred that the following classes were relevant: crane, bulldozer, cement mixer, construction sign, scaffolding, excavator, hard hat, construction worker, truck, and barrier.
The \gls{uav} incrementally built a semantic graph and provided it to the \gls{ugv}.
Throughout the mission, the UGV reported relevant findings to the human operator, who would then retask the \gls{ugv} based on those findings.
Overall, the \gls{ugv} traveled 756 meters and received 21 updates from the operator.
The UGV spent nearly 90\% of its time in autonomous mode.
The manual takeovers came primarily from dynamic obstacles, such as buses or trucks, or small obstacles that the \gls{ugv}\ obstacle detection could not identify, such as puddles or potholes.
The ratio of API calls to user interactions is lower than in the UGV-only experiments (See Table~\ref{tab:ugv_outcomes}).
This is because, in the \gls{ugv}-only experiments, the ground robot inferred all decisions autonomously, whereas in this experiment the \gls{ugv} would reach back to the human operator.
Throughout the mission, the UGV provides this information via detected objects and scene captions associated with the graph.
Importantly, the UGV also provides negative information (\eg ``this is a parking lot with no construction activity'').

\medskip
\noindent\textbf{System Demonstration 2: Triaging}
The second mission was specified by the query ``you are working with a high-altitude UAV to search for people.''
During mission execution, the user provided additional details.
For example, the user first instructed the UGV to look 30 south and 50 meters east, in order to complement the UAV's flight path.
SPINE infers the relevant classes are person, tree, car, bicycle, bench, backpack, streetlight, trash can, and umbrella.
Following a similar mission concept, the \gls{uav} is provided with the same list of classes to generate and provide a semantic map.
SPINE also provides an interpretable explanation for the chosen classes:  For instance, `Bench', `backpack', and `trash can' indicate places where people might be found or have left their belongings. `Buildings' and `streetlights' provide ``structural context to the environment''.
The \gls{ugv} works with the user to explore the environment.
Throughout the mission, the UGV provides natural language description of the previously unexplored area, which provides  situational awareness to the user and aids in planning.
Snapshots of the experiment are shown in Fig.~\ref{fig:ugv_mission_5}.
Overall, the UGV correctly identified five people.
The detection system made one false positive, and one person was in the UGVs field of view, but the tracker failed to register that person.

\begin{figure}[t]
    \centering
    \includegraphics[width=0.95\linewidth]{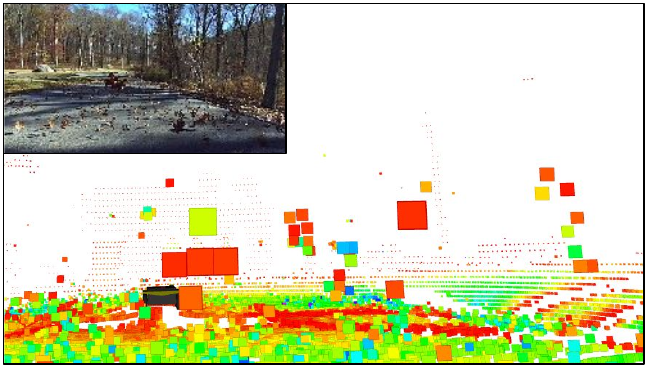}
    \caption{Example of odometry failure. A gust of leaves blew around the ground vehicles LiDAR. The spurious returns caused unrecoverable odometry drift. }
    \label{fig:ugv_odom_fail}
\end{figure}

\begin{table*}[h!]
    \centering
    \begin{tabular}{p{3.5cm}cccccc} \toprule
         Specification   &  UGV Distance (m) & Time (s) & User Interactions & API calls & Removed Edges & (\%) Autonomous \\ \toprule
         (S6) I heard of construction  around the eastern roads. Can you check? & 765 & 2097 & 21 &  41 & 8 & 90 \\
         (S7) You are working with a high-altitude UAV to search for people. & 695 & 2390 & 19 & 37 & 15 & 93 \\
        \toprule
    \end{tabular}
    \caption{System demonstration results over two different specifications. Both missions require the air ground system to acquire information in the Pennovation environment, thus the resulting metrics are similar. }
    \label{tab:sys_outcomes}
\end{table*}

\section{DISCUSSION, CONCLUSION, AND FUTURE WORK}
We conclude the paper by discussing our results, summarizing our key contributions, and outlining promising directions for future work.

\subsection{Discussion}

\begin{figure}[b!]
    \centering
    \includegraphics[width=0.95\linewidth]{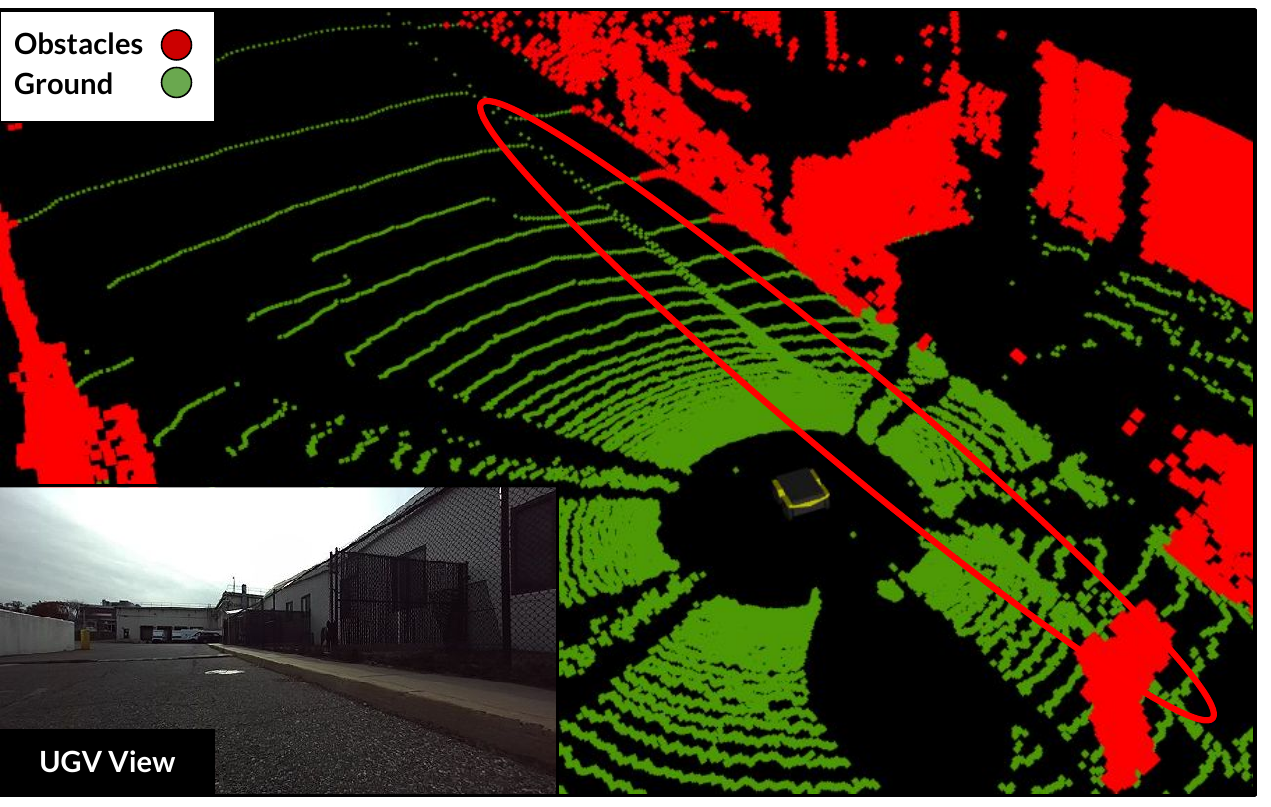}
    \caption{Example reason for manual takeover. Curbs (circled) are not detected by traversability estimation, thus UGV tries to drive over them.}
    \label{fig:ugv_curb}
\end{figure}

UGV traversability estimation was a primary system bottleneck.
The UGV struggled to identify negative and small positive obstacles, limiting the area that the UGV could safely explore. 
For example, Fig.~\ref{fig:ugv_curb} shows a portion from the second system demonstration where the UGV failed to identify a curb.
While the semantic graph from the \gls{uav} provided valuable traversability information, it contained false positives such as buildings marked as roads.
The UGV often identified these errors online and corrected its path.
However, the safety operator had to take over when false positives brought the UGV through curbs or other challenging obstacles. 
This is particularly apparent in Sec.~\ref{sec:sys_demo} where the specificity of the task is lower and a smaller model is used to generate the object masks leading to harder trajectories for the \gls{ugv} to follow leading to less time spent in autonomous mode.
The system used \gls{gnss} to register maps between robots, but the UGV used LiDAR odometry only for state estimation. 
While LiDAR odometry is typically a robust state estimation solution, it suffers from drift in highly dynamic scenes, as shown in  Fig.~\ref{fig:ugv_odom_fail}. 
Finally, we observed that the LLM-enabled planner was unable perform robust exploration, as evidenced by our UGV-only exploration experiments (see Fig~\ref{fig:ugv-exploration}), which made the UGV dependent on a strong prior from the UAV.

\begin{figure}
    \centering
    \includegraphics[width=0.95\linewidth]{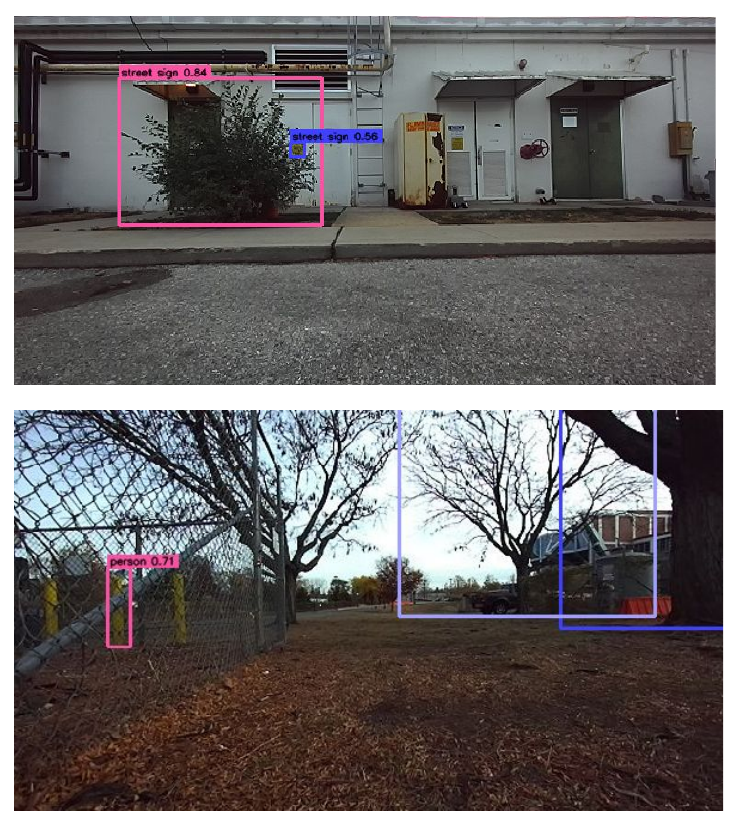}
    \caption{Example misclassifications made by open-vocabulary object detection. Depending on the mission specification, misclassifications can be benign (falsely detected street sign, top) or detrimental to the mission (falsely detected person, bottom).}
    \label{fig:enter-label}
\end{figure}

\subsection{Conclusion}
\label{sec:conclusion}

In summary, we present an air-ground teaming system for natural-language missions in unknown environments.
The system infers relevant semantics given a specification.
An aerial vehicle then incrementally builds a mission-relevant semantic map, which is relayed to a ground vehicle. 
The ground vehicle then uses an LLM-enabled planner to infer and realize subtasks that realize the mission, and the planner leverages online semantic mapping to augment and correct the  semantic map received from the aerial vehicle.
During the mission, the aerial vehicle intermittently provides a map update to the ground vehicle via an opportunistic communication network. 
We evaluate our system over seven different specifications in urban and rural environments in up to kilometer-scale navigation missions.

\subsection{Future Work \& Open Challenges}
We see several interesting directions for future work.
This work uses LLMs to translate and act upon instructions in natural language and leverages a structured map format for in-context updates.
While this empirically works, we lose information in the construction of the maps and are explicitly reliant on an uplink to communicate with the language model API.

\noindent \textbf{Onboard Language Models. } In this work, we use the OpenAI GPT 4o model, which makes us reliant on a robust communication infrastructure. 
In our ideal system, we would be able to decompose the main task from the user into subtasks that onboard LLMs can handle.
Keeping the interface between the robots to be natural language allows us to maintain the sparsity of the communication 
between the robots and interpretable by humans in the loop.

\noindent \textbf{Traversability Estimation. }
While segmenting aerial images appears to have good zero-shot performance, imperfections in the detections can lead to catastrophic failure in a \gls{ugv} with a naive local planner.
Failure cases such as minor localization errors, inability to perceive thin objects like fences, and the change in depth between roads and curbs may result in collision for a naive \gls{ugv} planner.
Additionally, while a coarse traversability estimation can be computed by segmenting roads, paths, and grass, these are not always meaningful for all robot modalities. 
For example, smaller robots will have difficulty traversing potholes, speed bumps, or curbs, which may be included in a road segmentation. 
Incorporating a robot-based traversability would be a valuable addition to this work. For instance,
in addition to the pre-defined \verb|road| label, we can task the aerial robot to find semantic classes that align with the robot's experience or claimed capability to paths better suited to the robot. 

Finally, in future work we plan to use dense depth images to find smaller objects like tree roots or potholes that robots can avoid based on their capabilities. 
In this framework, we can also imagine re-tasking of \glspl{uav} to regions in the environment that are semantically meaningful. 
Our pipeline is able to accommodate the geometric clustering of semantically meaningful objects into regions.
This in turn could be used to interactively ask the aerial robot to survey desired locations in the environment with tasks such as ``Map all the \verb|cars| in \verb|parklot_01|".

\section*{ACKNOWLEDGMENTS}
We would like to acknowledge 
Katie Mao, Frank Gonzalez, Dexter Ong, and Kashish Garg for their help during the field experiments for this work. We would like to acknowledge Dr. Michael Novitzky and Tyler Errico for their help and support during field experiments at USMA West Point.

\bibliographystyle{IEEEtran}
\bibliography{refs}

\clearpage
\section*{APPENDIX}
\renewcommand{\thesection} {A.\arabic{section}} 
\setcounter{section}{0}
\setcounter{subsection}{0}
\setcounter{figure}{0}
\renewcommand\thefigure{A.\arabic{figure}} 

\section{Platforms}
\begin{figure}[t]
    \centering
\includegraphics[width=.8\linewidth]{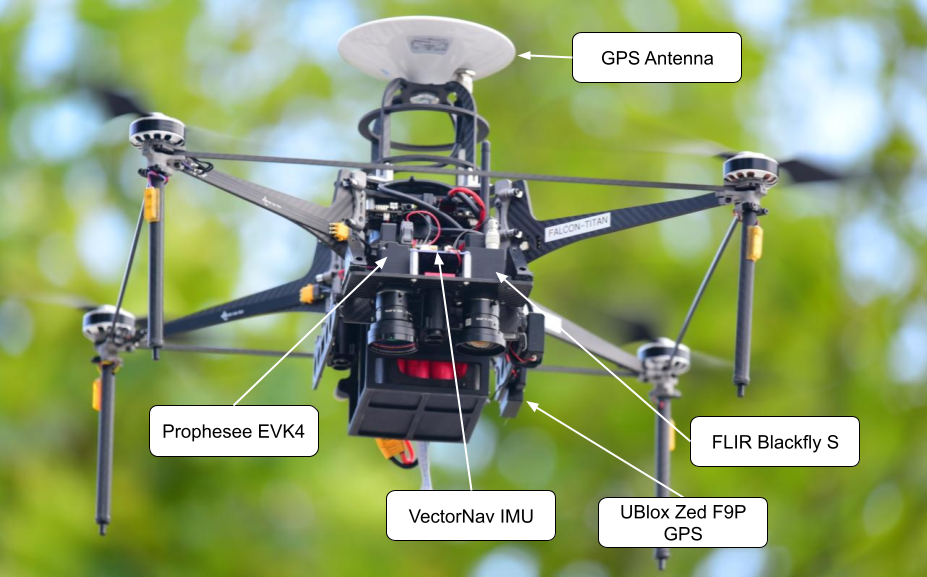}
    \caption{Falcon 4 platform configured for high-altitude semantic mapping missions.
    }
    \label{fig:app-falcon4}
\end{figure}

\begin{figure}[t]
    \centering
    \includegraphics[width=.8\linewidth]{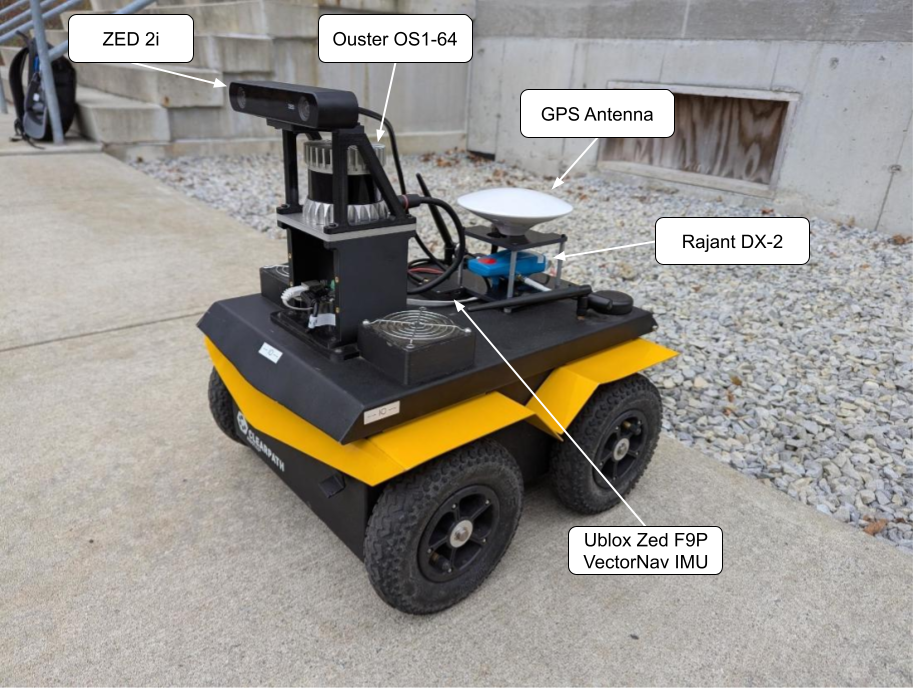}
    \caption{UGV platform used for ground experiments.
    }
    \label{fig:app-jackal}
\end{figure}

\begin{figure}[t!]
    \centering
    \includegraphics[width=\linewidth]{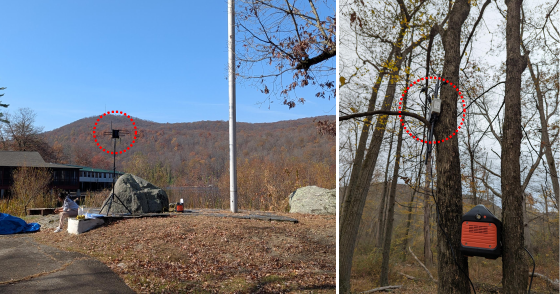}
    \caption{Communication nodes used for field experiments.
    \uline{Left}: FE1 base station node.
    \uline{Right}: \emph{dummy} ME4 node used for LLM API calls. }
    \label{fig:comm-nodes}
\end{figure}

While our methodology can support other platforms, we use a custom built quadcopter and a ClearPath Jackal for this work.
Figs.~\ref{fig:app-falcon4} and \ref{fig:app-jackal} shows the platforms used during our experiments.
Both platforms are suited with compute capabilities to generate the semantic mapper onboard.
The choice of compute onboard these platforms is the result of several iterations of field experiments and are targeted towards the maximum available CPU \& GPU compute for the power budgets for the respective platforms.
The power budget allowed for the Jackal is 200W and the \gls{uav} is allowed a power budget of 100W.
The \gls{ugv} and \gls{uav} both have a battery life of approximately 30 minutes.
Both platforms were fitted with Rajant mesh radios. We also used \emph{dummy} nodes (Fig.~\ref{fig:comm-nodes}) as communication relays for SPINE, as described in Sec.~III.\ref{subsec:communications}.


\section{Air router \& State machine}
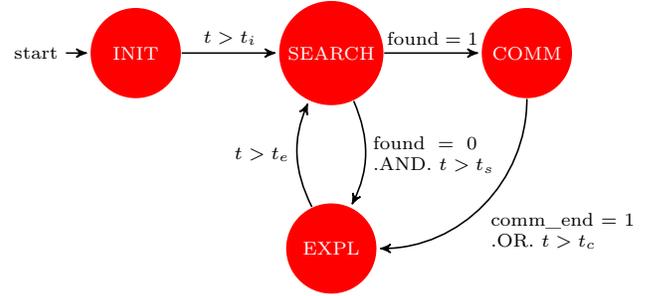
\begin{figure}
    \centering
    \begin{tikzpicture}[->,>=stealth',shorten >=1pt,auto,node distance=2.6cm,
                    semithick]
  \tikzstyle{every state}=[fill=red,draw=none,text=white, minimum size=1.2cm]
\scriptsize
  \node[initial,state] (A)              {INIT};
  \node[state]         (B) [right of=A] {SEARCH};
  \node[state]         (D) [right of=B] {COMM};
  \node[state]         (C) [below of=B] {EXPL};

  \path (A) edge              node {$t > t_i$} (B)
        (B) edge              node {$\text{found} = 1$} (D)
        (D) edge [bend left=45]  node [text width=2cm,align=left]{comm\_end = 1 .OR. $t>t_c$} (C)
        (B) edge [bend left=25]  node [text width=2cm,align=left]{$\text{found} = 0$ .AND. $t > t_s$} (C)
        (C) edge [bend left=25]  node {$t > t_e$} (B);
\end{tikzpicture}
    \vspace{-.2cm}
    \caption{UAV mission executed during the air-ground experiments.
    Figure from \cite{cladera2024enabling}. }
    \label{fig:uav_sm}
\end{figure}

We relied on \texttt{air\_router}~\cite{cladera2024challenges} to guide the \gls{uav} during the experiments. This high-level state machine, illustrated in Fig.~\ref{fig:uav_sm}, transitions the \gls{uav} between an exploration and a communication mission. It also looks for ground robots (based on the last known position or last known goal) to deliver messages if communication has not occurred recently.

\section{UAV Localization}
\label{sec:app_localization}
In order to localize objects within images accurately we need an extremely accurate pose estimate for when each image was taken.
To this end, we use GTSAM \cite{dellaert2012factor} to fuse GPS position estimates with IMU readings.
For our IMU we use a Vectornav V100T which runs its own Kalman Filter to fuse magnetometer with gyroscope and accelerometer readings. This gives us north aligned orientation data from the IMU at 400 hz.
We get 5 hz global position readings from the GPS which we model as unary pose factors \cite{dellaert2012factor} and use the preintegrated IMU measurements \cite{forster2016manifold} to constrain the pose graph.
As the UAV can cover several kilometers we use the iSAM2 \cite{kaess2012isam2} optimizer to maintain a reasonable latency over large distances.
We run the optimizer at each GPS measurement.
We ensure that we do not include erroneous GPS measurements by ensuring that they lie within 2$\sigma$ of the estimated pose and covariance.
We also predict the pose at 100 hz by integrating the IMU measurements from the last optimized pose to the current time.
We note that while we use this particular set of sensors in this work, the framework could be extended to include other sensors such as barometers (for height) and raw magnetometer readings directly.
This allows us to get sub one-tenth of a second pose estimates for each image.

\section{Active Perception}
\begin{figure}[t]
    \centering
    \includegraphics[width=0.95\linewidth]{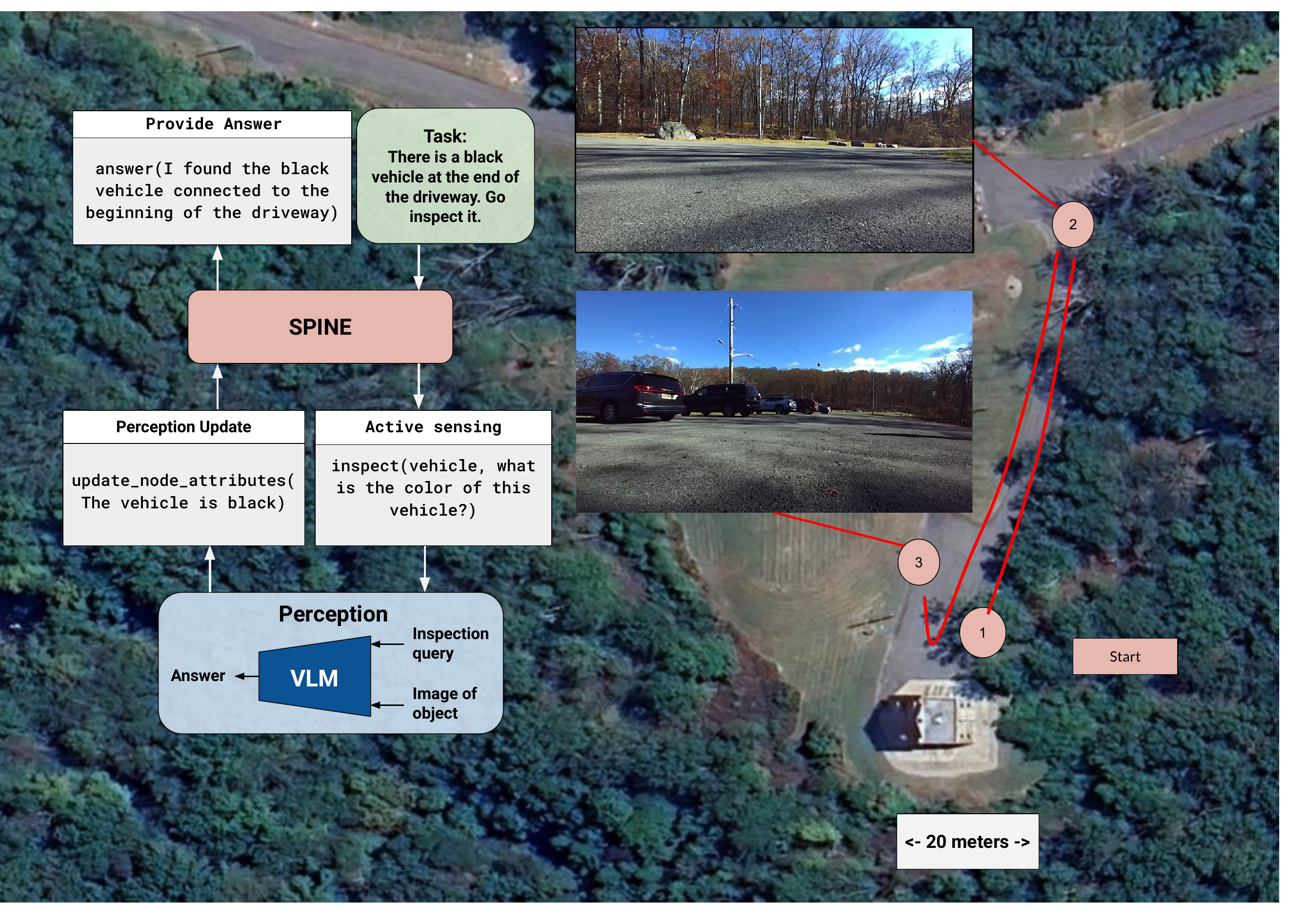}
    \caption{Example of SPINE using active perception to correct mission misspecification. User tasks ground vehicle with inspecting a black vehicle at the end of the driveway (1). However, there is no ground vehicle at the end of the driveway (2). The ground vehicle identifies several cars and trucks at the start of driveway, and inspects those. The ground vehicle forms a mission-relevant perception query ``is this vehicle black'' in order to resolve the mission.}
    \label{fig:vlm_misspec}
\end{figure}

SPINE uses active perception to resolve errors in the mission specification. Fig.~\ref{fig:vlm_misspec} shows a mission where the UGV is provided the task: ``There is a black vehicle at the end of the driveway. Go inspect it.'' The UGV goes to the end of the driveway but does not find a vehicle.
During navigation, it observes the vehicle at the beginning of the driveway.
So, it doubles back to observe those, and queries perception to find the color of the vehicle.

\begin{figure}[t]
    \centering
    \includegraphics[width=0.95\linewidth]{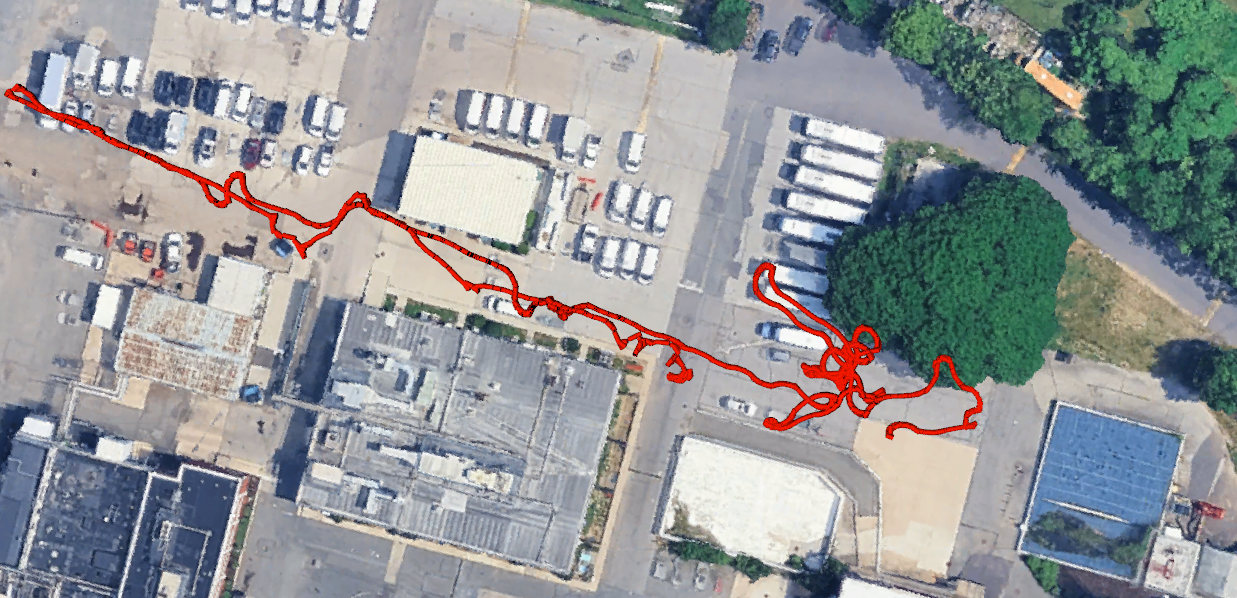}
    \caption{UGV trajectory from the first system demonstration. Using the UAV-generated graph, the UAV travels over 700 meters while finding objects and describing regions of the enviroment.}
    \label{fig:ugv_sys_demo}
\end{figure}

\section{Large-scale demonstrations}

We plot the UAV path from the first system demonstration, as reported in Sec~\ref{sec:sys_demo}, in Fig.~\ref{fig:ugv_sys_demo}.
Please note that the trajectory is overlayed on a Google Earth image, so the vehicles displayed were not present in the actual experiment.

\end{document}